\documentclass{article}
\PassOptionsToPackage{numbers,compress,sort}{natbib}
\usepackage[final]{neurips_data_2021}

\usepackage[subfig]{definition}

\usepackage{tcolorbox}
\tcbuselibrary{skins}
\tcbset{skin=widget,left=0.2em,right=0.2em,top=0.2em,bottom=0.2em}

\newcommand{\thesystem}{\textsf{PyGCL}\xspace}

\title{An Empirical Study of Graph Contrastive Learning}

\author{%
Yanqiao Zhu\textsuperscript{\textmd{1,2}}
\qquad
Yichen Xu\textsuperscript{\textmd{3}}
\qquad
Qiang Liu\textsuperscript{\textmd{1,2}}
\qquad
Shu Wu\textsuperscript{\textmd{1,2}}\thanks{To whom correspondence should be addressed.} \\
\textsuperscript{1}Center for Research on Intelligent Perception and Computing\\ Institute of Automation, Chinese Academy of Sciences\\
\textsuperscript{2}School of Artificial Intelligence, University of Chinese Academy of Sciences\\
\textsuperscript{3}School of Computer Science, Beijing University of Posts and Telecommunications\\
\small\texttt{yanqiao.zhu@cripac.ia.ac.cn\quad linyxus@bupt.edu.cn\quad }
\small\texttt{\{qiang.liu,shu.wu\}@nlpr.ia.ac.cn} \\
}

\begin{document}

\maketitle

\begin{abstract}
Graph Contrastive Learning (GCL) establishes a new paradigm for learning graph representations without human annotations.
Although remarkable progress has been witnessed recently, the success behind GCL is still left somewhat mysterious.
In this work, we first identify several critical design considerations within a general GCL paradigm, including augmentation functions, contrasting modes, contrastive objectives, and negative mining techniques.
Then, to understand the interplay of different GCL components, we conduct extensive, controlled experiments over a set of benchmark tasks on datasets across various domains.
Our empirical studies suggest a set of general receipts for effective GCL, e.g., simple topology augmentations that produce sparse graph views bring promising performance improvements; contrasting modes should be aligned with the granularities of end tasks.
In addition, to foster future research and ease the implementation of GCL algorithms, we develop an easy-to-use library \thesystem, featuring modularized CL components, standardized evaluation, and experiment management.
We envision this work to provide useful empirical evidence of effective GCL algorithms and offer several insights for future research.
\end{abstract}

\section{Introduction}

The past few years have witnessed rapid advances in Graph Neural Networks (GNNs) \cite{Kipf:2016tc,Velickovic:2018we}, which is a powerful framework for analyzing graph-structured data. As the most GNN models focus on (semi-)supervised learning, which requires access to abundant labels, recent trends in Self-Supervised Learning (SSL) have led to a proliferation of studies that learns graph representations without relying on human annotations.
Among SSL methods, Contrastive Learning (CL), also known as instance discrimination, is a major area of interest and has already achieved comparable performance with its supervised counterparts in many representation learning tasks \cite{Hjelm:2019uk,Bachman:2019wp,He:2020tu,Caron:2020uv,Grill:2020uc,Chen:2020wj,Qian:2021tp,Klein:2020jb,Fang:2020ux,Gao:2021wf}.

Recently, remarkable progress has been made to adapt CL techniques for the graph domain. 
A typical Graph Contrastive Learning (GCL) method constructs multiple graph views via stochastic augmentations of the input at first and then learns representations by contrasting positive samples against negative ones.
For each node being an anchor instance, its positive samples are often chosen as the congruent representations in other views, while negatives are selected from other nodes within the given graph or other graphs within the batch.
Although GCL has constituted a new paradigm of SSL in the graph domain and achieved promising results, recent studies \cite{Velickovic:2019tu,Hassani:2020un,Peng:2020gw,You:2020ut,Zhu:2020vf,Zhu:2021wh,Sun:2020vi} seem to resemble each other with very limited nuances from the methodological perspective.
Moreover, most existing work only provides model-level evaluation and the contributing factors leading to the success of GCL still remain somewhat mysterious, which calls for a deeper understanding of different GCL components.

Towards this end, we try to shed light on how these GCL algorithms succeeded through the lens of empirical evaluation of critical design considerations.
We first propose a general contrastive paradigm which characterizes previous work by limiting the design space of interest to four dimensions: (a) data augmentation functions, (b) contrasting modes, (c) contrastive objectives, and (d) negative mining strategies.
Note that we include no model-specific design considerations such as the number of attention heads for graph attentive encoders. To the best that we are aware, these four dimensions cover a wide range of options that are representative in open literature.

Then, we systematically study the empirical performance of different design dimensions through controlled experiments over benchmark tasks on a set of datasets across a variety of domains.
With the empirical studies, we attempt to provide answers to the following questions:
\begin{itemize}[leftmargin=*]
	\item What is the most contributory component in an effective GCL algorithm? 
	\item How do different design considerations affect the model performance?
	\item Do these design considerations favor certain types of data or end tasks?
\end{itemize}
We note that there has been several survey papers on self-supervised graph representation learning \cite{Xie:2021uv,Liu:2021wc,Wu:2021wg}.
However, to the best of our knowledge, none of existing work provides rigorous empirical evidence on the impact of each component in GCL.

We summarize several key findings of the empirical study, which we hope could benefit the graph SSL community for developing future algorithms.
Our experiments suggest a set of general recipes for effective GCL algorithms:
\begin{itemize}[leftmargin=*]
	\item GCL algorithms benefit the most from topology augmentations that produce sparse graph views. In addition, bi-level augmentation on both topology and feature levels further improves the performance.
	\item Overall, same-scale contrasting modes are desirable. The contrasting modes should also be chosen according to the granularity of downstream tasks.
	\item The InfoNCE objective obtains stable, consistent performance improvements under all settings yet requires a large number of negative samples.
	\item Several recently proposed negative-sample-free objectives have great potential for reducing computational burden without compromising on performance.
	\item Current negative mining strategies based on calculating embedding similarities bring limited performance improvements to GCL.	
\end{itemize}

In addition, to foster future research, we develop \thesystem, an easy-to-use PyTorch framework, featuring commonly used, modularized GCL components, standardized evaluation, and experiment management. We hope the use of \thesystem will greatly relief the burden of comparing existing baselines and developing new algorithms. The \thesystem is open-sourced at \url{https://github.com/GraphCL/PyGCL}.

\section{A General Paradigm of GCL and its Design Dimensions}

\textbf{Problem formulation.}
Let \(\mathcal{G} = (\mathcal{V}, \mathcal{E})\) denote a given graph, where \(\mathcal{V} = \{v_i\}_{i=1}^N\), \(\mathcal{E} \subseteq \mathcal{V} \times \mathcal{V}\) represent the node set and the edge set respectively. We further denote the feature matrix and the adjacency matrix as \(\bm{X} \in \mathbb{R}^{N \times F}\) and \(\bm{A} \in \{0,1\}^{N \times N}\), where \(\bm{x}_i \in \mathbb{R}^{F}\) is the feature of \(v_i\) and \(\bm{A}_{ij} = 1\) iff \((v_i, v_j) \in \mathcal{E}\).
In the setting of unsupervised representation learning, there is no given class information of nodes or graphs during training.
Our objective is to learn a GNN encoder \(f(\cdot)\) receiving the graph features and structure as input, that produces node embeddings in low dimensionality. We denote \(\bm{H} = f(\bm{X}, \bm{A}) \in \mathbb{R}^{N \times F^\prime}\) as the learned representations of nodes, where \(\bm{h}_i\) is the embedding of node \(v_i\). For graph-oriented tasks, we can further obtain a graph-level representation \(\bm{s} = r(\bm{H}) \in \mathbb R^{F^\prime}\) of \(\mathcal{G}\) that aggregates node-level embeddings. Note that the readout function \(r(\cdot)\) might be a simple permutation-invariant function such as mean or sum pooling, or may be parameterized by a neural network. These representations can be used in downstream tasks, such as node/graph classification and community detection.

\begin{figure*}
	\centering
	\includegraphics[width=0.75\linewidth]{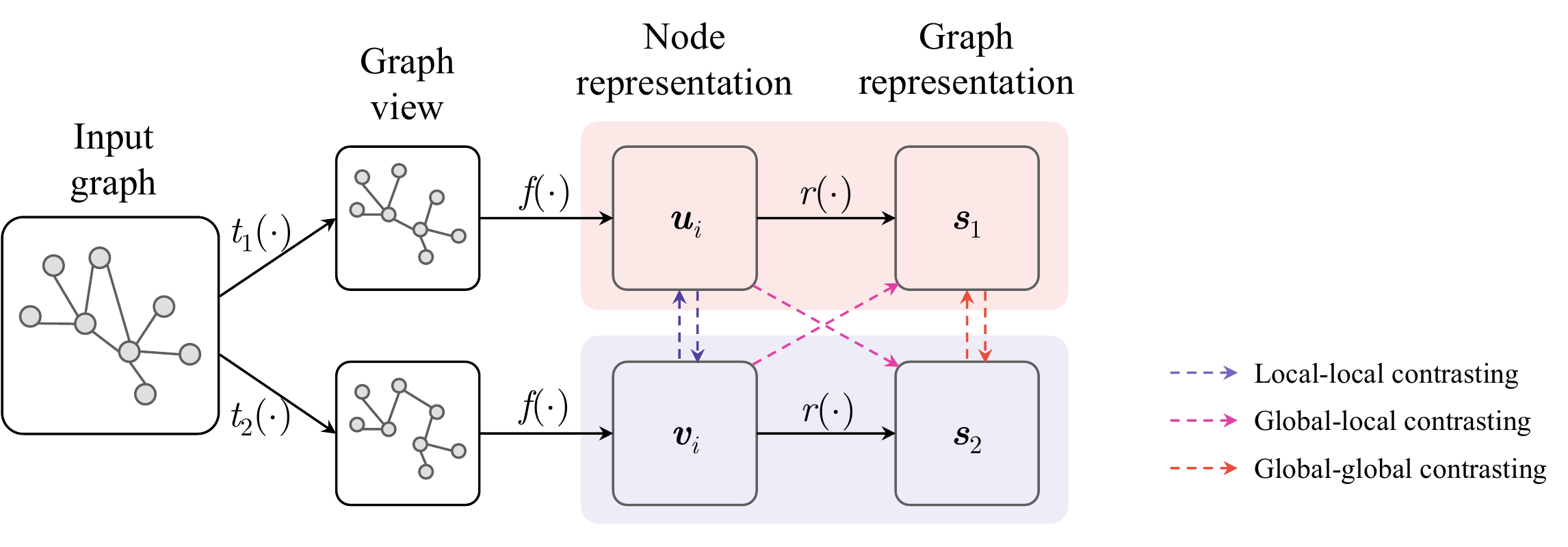}
	\caption{A general GCL model. At first, two graph views are generated via data augmentation functions. Then, the two graphs are fed into a shared graph neural network to learn representations, which are then optimized with a contrastive objective that pulls congruent representation pairs in the two views together while pushing others away. Additional negative mining techniques may be employed to improve the model performance.}
	\label{fig:model}
\end{figure*}

\textbf{A general paradigm of GCL.}
We decompose GCL algorithms from four dimensions: (a) data augmentation functions, (b) contrastive mode, (c) contrastive objective, and (d) negative mining strategies. These four components constitute the design space of interest in this work.

At each iteration of training, we first perform stochastic \textbf{augmentations} to generate multiple graph views from the input graph.
Specifically, we sample two augmentation functions \(t_1, t_2 \sim \mathcal{T}\) to generate graph views \(\widetilde{\mathcal{G}}_1 = t_1(\mathcal{G})\) and \(\widetilde{\mathcal{G}}_2 = t_2(\mathcal{G})\), where \(\mathcal{T}\) is the set of all possible transformation functions to be discussed in the next section.
We then obtain node representations for the two views using a shared GNN encoder \(f(\cdot)\), denoted by \(\bm{U} = f(\widetilde{\bm{X}}_1, \widetilde{\bm{A}}_1)\) and \(\bm{V} = f(\widetilde{\bm{X}}_2, \widetilde{\bm{A}}_2)\) respectively. Optionally, we may obtain graph representations for each graph view through a readout function \(r(\cdot)\): \(\bm{s}_1 = r(\bm{U})\) and \(\bm{s}_2 = r(\bm{V})\).

For every node embedding \(\bm{v}_i\) being the anchor instance, the \textbf{contrasting mode} specifies a positive set \(\mathcal{P}(\bm{v}_i) = \{\bm{p}_i\}_{i=1}^P\) and a negative set \(\mathcal{Q}(\bm{v}_i) = \{\bm{q}_i\}_{i=1}^Q\).
In a pure unsupervised learning setting, we only consider congruent samples in each graph view; in other words, embeddings in the two augmented graph views corresponding to the same node or graph constitute the positive set.
It is noted that when label supervision is given, the positive set may be enlarged with samples belonging to the same class \cite{Khosla:2020tr}.
Moreover, we may employ \textbf{negative mining strategies} to improve the negative sample set by considering the relative similarity (i.e. the hardness) of negative samples.
Finally, we use a \textbf{contrastive objective} \(\mathcal{J}\) to score these specified positive and negative pairs.%

In the following, we succinctly list implementations of these four design dimensions considered in this work. For details of each implementation, we refer readers of interest to Appendix \ref{sec:design-dimensions-details}.

\subsection{Design Dimensions}

\textbf{Data augmentations.}
The purpose of data augmentation is to generate \emph{congruent, identity-preserving positive samples} of the given graph. Most GCL work involves bi-level augmentation techniques: topology (structure) transformation and feature transformation.
\begin{itemize}[leftmargin=*]
	\item \textbf{Topology augmentations:} (1) Edge Removing (ER), (2) Edge Adding (EA), (3) Edge Flipping (EF), (4) Node Dropping (ND), (5) Subgraph induced by Random Walks (RWS), (6) diffusion with Personalized PageRank (PPR), and (7) diffusion with Markov Diffusion Kernels (MDK).
	\item \textbf{Feature augmentations:} (1) Feature Masking (FM) and (2) Feature Dropout (FD).
\end{itemize}

\textbf{Contrasting modes.}
For an anchor instance, contrasting modes determine the positive and negative sets at different granularities of the graph.
In mainstream work, three contrasting modes are widely employed: (1) local-local CL and (2) global-global CL, which contrast embeddings at the same scale, and (3) global-local CL, which contrasts cross-scale embeddings.
Note that the space of contrasting mode depends on downstream tasks. Only local-local and global-local CL are applicable for node datasets, where all three mode can be used for graph datasets.

\textbf{Contrastive objectives.}
Contrastive objectives are used to measure the similarity of positive samples and the discrepancy between negatives. We consider the following objective functions that rely on negative samples: (1) Information Noice Contrastive Estimation (InfoNCE), (2) Jensen-Shannon Divergence (JSD), and (3) Triplet Margin loss (TM).
Moreover, we also analyze the following objectives that eschew the need of explicit negative samples: (4) the Bootstrapping Latent loss (BL), (5) Barlow Twins (BT) loss, and (6) VICReg loss.

\textbf{Negative mining strategies.}
Recent work argues that CL algorithms benefits from hard negative samples (i.e. samples difficult to distinguish from an anchor instance).
In this work, we consider the following four negative mining strategies: (1) Hard Negative Mixing (HNM), (2) Debiased Contrastive Learning (DCL), (3) Hardness-Biased Negative Mining (HBNM), and (4) Conditional Negative Mining (CNM).

\begin{table*}[t]
	\centering
	\caption{Summary of representative GCL models within the proposed paradigm.}
	\label{tab:summary}
	\resizebox{\linewidth}{!}{
	\begin{tabular}{ccccccc}
	\toprule
	Method & Primary task & \makecell{Topology\\augmentation} & \makecell{Feature\\augmentation} & \makecell{Contrasting\\mode} & \makecell{Dual\\branches?} & \makecell{Contrastive\\objective} \\ %
	\midrule
	DGI \cite{Velickovic:2019tu} & Node classification & --- & --- & Global-local & \xmark & JSD \\
	GMI \cite{Peng:2020gw} & Node classification & --- & --- & Global-local & \xmark & SP-JSD \\
	InfoGraph \cite{Sun:2020vi} & Graph classification & --- & --- & Global-local & \xmark & SP-JSD \\
	MVGRL \cite{Hassani:2020un} & Node \& graph classification & PPR & --- & Global-local & \cmark & JSD \\
	GCC \cite{Qiu:2020gq} & Transfer learning & RWS & --- & Local-local & \xmark & InfoNCE \\
	GraphCL \cite{You:2020ut} & Graph classification & RWS/ND/EA/ED & FD & Global-global & \cmark & InfoNCE \\
	GRACE \cite{Zhu:2020vf} & Node classification & ER & MF & Local-local & \cmark & InfoNCE \\
	GCA \cite{Zhu:2021wh} & Node classification & ER & MF & Local-local & \cmark & InfoNCE \\
	BGRL \cite{Thakoor:2021tl} & Node classification & ER & MF & Local-local & \cmark & BL \\
	GBT \cite{Bielak:2021uv} & Node classification & ER & MF & Local-local & \cmark & BT \\
	\bottomrule
	\end{tabular}
	}
\end{table*}

\subsection{Discussions on Representative GCL Methods}
We give a brief summary of existing representative GCL methods as shown in Table \ref{tab:summary} and discuss how they fit into our proposed paradigm.
We note that negative mining strategies have received scant attention in current GCL literature and thus are omitted in the table.

\textbf{Dual branches vs. single branch.}
We notice that most work leverages a dual-branch architecture following SimCLR \cite{Chen:2020wj} that augments the original graph twice to form two views and designates positive samples across two views.
For some global-local CL methods like DGI \cite{Velickovic:2019tu} and GMI \cite{Peng:2020gw}, they employ an architecture with only one branch. In this case, negative samples are obtained by corrupting the original graph. Different from the aforementioned \emph{augmentation} schemes that generate congruent pairs to model \emph{the joint distribution} of positive pairs, we resort to the term \emph{corruption functions}, which approximate \emph{the product of marginals}.

\textbf{Stronger augmentations.}
Unlike GRACE \cite{Zhu:2020vf} and GraphCL \cite{You:2020ut} that employ uniform edge/feature perturbation, GCA \cite{Zhu:2021wh} proposes to perform adaptive augmentation based on importance scores of edges and features. In this work, to involve less hyperparameters as possible, we focus on uniform transformation only.

\textbf{Variants of contrasting modes.}
GMI \cite{Peng:2020gw} extends DGI \cite{Velickovic:2019tu} by further considering the agreement between raw node/edge features and node/edge representations. Because it requires much more computational resources, our experiments exclude this implementation of contrasting mode.
In addition, there are several recent methods \cite{Zhang:2020ts,Mavromatis:2021jr} involve contrasting between local/global and \emph{context} representations, which are usually derived from graph clustering algorithms. Considering the generality of the experiments, we shall leave studying these variants as a future direction.

\section{Empirical Studies}

The following section presents the empirical studies of GCL components.
In the following section, we first introduce experimental configurations and then summarize the results and observations regarding each particular component in the proposed paradigm.
We refer interested readers to Appendix \ref{sec:experimental-protocols} for more information on the evaluation protocols and implementations.

\textbf{Evaluation configurations.}
We conduct experiments on a variety of medium- to large-scale datasets widely used in literature, ranging from academic networks to chemistry molecules. For fair comparison, we closely follow previous studies on datasets preprocessing \cite{Velickovic:2019tu,You:2020ut,Zhu:2020vf,Zhu:2021wh,Sun:2020vi,Thakoor:2021tl,Shchur:2018vv,Morris:2020wd}.
We mainly evaluate models with different design considerations on two essential tasks: (1) unsupervised node classification and (2) unsupervised graph classification.
For all experiments, we follow the linear evaluation scheme used by DGI \cite{Velickovic:2018we}. Particularly, the models are first trained in an unsupervised manner and the resulting embeddings are fed into a linear classifier to fit the labeled data.
We run the model with ten random splits and report the averaged accuracies (\%) as well as the standard deviation.

\begin{table}
	\centering
	\caption{Statistics of datasets used for unsupervised learning experiments.}
	\vskip 0.5em
    \label{tab:dataset}
	\resizebox{\linewidth}{!}{
	\begin{tabular}{cccccccc}
    \toprule
    Dataset & Domain & Task  & \#Graphs & Avg. \#nodes & Avg. \#edges & \#Features & \#Classes \\
    \midrule
    Wiki & Knowledge base & \multirow{4}[0]{*}{\makecell{Unsupervised\\ node \\ classification}} & 1     & 11,701 & 216,123 & 300   & 10 \\
    Computer & Social networks &       & 1     & 13,752 & 245,861 & 767   & 10 \\
    CS & Citation networks &       & 1     & 18,333 & 81,894 & 6,805 & 15 \\
    Physics & Citation networks &       & 1     & 34,493 & 247,962 & 8,415 & 5 \\
    \midrule
    NCI1 & Biochemical molecules  & \multirow{4}[0]{*}{\makecell{Unsupervised\\ graph \\ classification}} & 4110  & 29.87 & 32.30 & --- & 2 \\
    PROTEINS & Bioinformatics &       & 1,133 & 39.06 & 72.82	 & 29 & 2 \\
    IMDB-M & Social networks &       & 1,500 & 13.00 & 65.94 & ---   & 3 \\
    COLLAB & Social networks &       & 5,000 & 74.49 & 2457.78 & --- & 3 \\
    \bottomrule
    \end{tabular}
    }
\end{table}

\textbf{Datasets.}
We conduct experiments on a variety of medium-scale datasets widely used in open literature, ranging from academic networks to chemistry molecular datasets, including Wiki-CS (Wiki) \cite{Mernyei:2020wh}, Amazon-Computer (Computer), Coauthor-CS (CS), and Coauthor-Physics (Physics) \cite{Shchur:2018vv} for node classification and NCI1 \cite{Wale:2006kv}, PROTEINS-full (PROTEINS) \cite{Borgwardt:2005jk}, IMDB-MULTI (IMDB-M), and COLLAB \cite{Yanardag:2015fm} for graph classification.
The statistics of all datasets is summarized in Table \ref{tab:dataset}.

\textbf{Implementation details.}
We choose Graph Convolutional Networks (GCNs) \cite{Kipf:2016tc} as the encoder for all node tasks and use Graph Isomorphism Networks (GINs) \cite{Xu:2019ty} for all graph tasks. For projector functions, we utilize an additional MultiLayer Perceptron (MLP) model.
To ensure convincing experiments and observations, we first perform an exhaustive search over the entire design space. Then, we select and report representative results to reveal common, useful practices.
With the aim of conducting controlled experiments, we fix as many variables, e.g., the GNN encoder architecture, embedding dimensions, number of epochs, and activation functions, as possible for every dataset.

\subsection{Data Augmentations}

\begin{tcolorbox}
\resizebox{\linewidth}{!}{
Augmentations: \textit{eval} \quad Contrasting mode: InfoNCE \quad Objectives: L--L \quad Negative mining strategy: None
}
\end{tcolorbox}

We first investigate how data augmentations affect the performance of GCL.
Specifically, we apply different data augmentation functions to generate two views, leverage the InfoNCE objective, and contrast local-local (node) representations. Except for specific augmentation functions used, all other settings are kept the same.
We first employ different topology and feature augmentations for GCL.
Then, we examine the use of compositional data augmentation schemes: (1) structure- and feature-level augmentations and (2) deterministic plus stochastic augmentations.

\phantomsection\label{obs:topology-augmentation}
\textbf{Observation 1. Topology augmentation greatly affects model performance. Augmentation functions that produce sparser graphs generally lead to better performance.}

\begin{table}
	\centering
	\caption{Classification performance with different topology and feature augmentations. The best performance is highlighted in boldface and the second-to-best underlined. OOM indicates Out-Of-Memory on a 24GB GPU.}
	\vskip 0.5em
	\resizebox{\linewidth}{!}{
	\begin{tabular}{cccccccccc}
	\toprule
	& \multirow{2.5}[0]{*}{Aug.} & \multicolumn{4}{c}{Node} & \multicolumn{4}{c}{Graph} \\
	\cmidrule(lr){3-6} \cmidrule(lr){7-10}
	& & Wiki  & CS    & Physics & Computer & NCI1 & PROTEINS & IMDB-M & COLLAB \\
	\midrule
	& None & 68.52{\footnotesize ±0.39} & 90.76{\footnotesize ±0.05} & 93.69{\footnotesize ±0.73} & 80.62{\footnotesize ±0.62} & 58.49{\footnotesize ±2.21} & 70.94{\footnotesize ±1.13} & 45.07{\footnotesize ±1.70} & 66.21{\footnotesize ±0.92} \\
	\midrule
	\parbox[t]{0mm}{\multirow{7}{*}{\rotatebox[origin=c]{90}{\small Topo.}}}
	&    EA    & 72.65{\footnotesize ±0.43} & 92.73{\footnotesize ±0.10} & 94.77{\footnotesize ±0.05} & 83.40{\footnotesize ±0.64} & 70.80{\footnotesize ±0.55} & 71.17{\footnotesize ±0.63} & 44.80{\footnotesize ±1.43} & 68.12{\footnotesize ±0.63} \\
    &    ER    & 76.38{\footnotesize ±0.21} & 92.83{\footnotesize ±0.17} & \underline{95.21{\footnotesize ±0.05}} & \textbf{87.84{\footnotesize ±0.76}} & 73.03{\footnotesize ±0.48} & \underline{72.55{\footnotesize ±0.11}} & 45.17{\footnotesize ±1.64} & 68.13{\footnotesize ±0.82} \\
    &    EF    & 74.10{\footnotesize ±0.67} & \underline{92.99{\footnotesize ±0.15}} & 94.88{\footnotesize ±0.06} & 86.68{\footnotesize ±0.73} & \underline{73.95{\footnotesize ±0.49}} & 70.64{\footnotesize ±1.67} & 44.15{\footnotesize ±1.21} & 67.92{\footnotesize ±0.93} \\
    &    ND    & \textbf{77.47{\footnotesize ±0.32}} & 92.81{\footnotesize ±0.08} & \textbf{95.99{\footnotesize ±0.12}} & 87.01{\footnotesize ±0.54} & 72.12{\footnotesize ±1.38} & \textbf{72.54{\footnotesize ±0.43}} & \textbf{47.03{\footnotesize ±1.14}} & \underline{70.73{\footnotesize ±0.78}} \\
    &    PPR   & 69.28{\footnotesize ±0.22} & 92.25{\footnotesize ±0.07} & OOM   & 85.06{\footnotesize ±0.53} & 58.70{\footnotesize ±0.51} & 71.69{\footnotesize ±1.12} & \underline{45.27{\footnotesize ±0.85}} & 68.51{\footnotesize ±0.67} \\
    &    MKD   & 69.87{\footnotesize ±0.12} & 92.62{\footnotesize ±0.14} & OOM   & 82.46{\footnotesize ±0.58} & 57.21{\footnotesize ±0.31} & 71.31{\footnotesize ±0.11} & 45.07{\footnotesize ±1.16} & 68.09{\footnotesize ±0.88} \\
    &    RWS   & \underline{76.74{\footnotesize ±0.20}} & \textbf{93.48{\footnotesize ±0.08}} & 95.04{\footnotesize ±0.11} & \underline{87.60{\footnotesize ±0.63}} & \textbf{75.11{\footnotesize ±1.14}} & 71.79{\footnotesize ±0.82} & 44.95{\footnotesize ±0.82} & \textbf{70.85{\footnotesize ±0.89}} \\
    \midrule
	\parbox[t]{0mm}{\multirow{2}{*}{\rotatebox[origin=c]{90}{\small Feat.}}}
	& FM    & \textbf{76.74{\footnotesize ±0.34}} & 91.55{\footnotesize ±0.11} & 94.12{\footnotesize ±0.21} & \textbf{85.05{\footnotesize ±0.51}} & \textbf{64.87{\footnotesize ±0.36}} & 71.35{\footnotesize ±0.79} & 45.36{\footnotesize ±1.68} & 70.52{\footnotesize ±0.35} \\
    & FD    & 76.68{\footnotesize ±0.16} & \textbf{91.83{\footnotesize ±0.08}} & \textbf{94.20{\footnotesize ±0.16}} & 84.93{\footnotesize ±0.46} & 63.21{\footnotesize ±0.51} & \textbf{71.60{\footnotesize ±1.61}} & \textbf{46.44{\footnotesize ±0.96}} & \textbf{70.69{\footnotesize ±1.33}} \\
	\bottomrule
	\end{tabular}
	}
	\label{tab:topology-feature-augmentations}
	\vskip -0.5em
\end{table}

Table \ref{tab:topology-feature-augmentations} summarizes the results of employing different topology and feature augmentations..
It is evident that the performance of GCL is highly dependent on the choice of topology augmentation functions.
We observe that models that remove edges (ER, MDK, ND, PPR, and RWS), compared to models that add edges (EA), in general achieve better performance, which suggests that augmentation functions produce \emph{sparser graph views} are preferable. %
We also find that RWS achieves better performance on node datasets, while ND favors graph tasks. We note that although random walk sampling is able to better extract local structural patterns, the graph datasets used in our study are of relatively small scales (< 500 nodes per graph). Therefore, these random-walk-based sampling strategies may be confined and a simple node dropping (ND) scheme generally outperforms other augmentations on graph-level tasks.

\begin{figure}
	\centering
	\subfloat[Node dropping probability]{
		\includegraphics[scale=0.53]{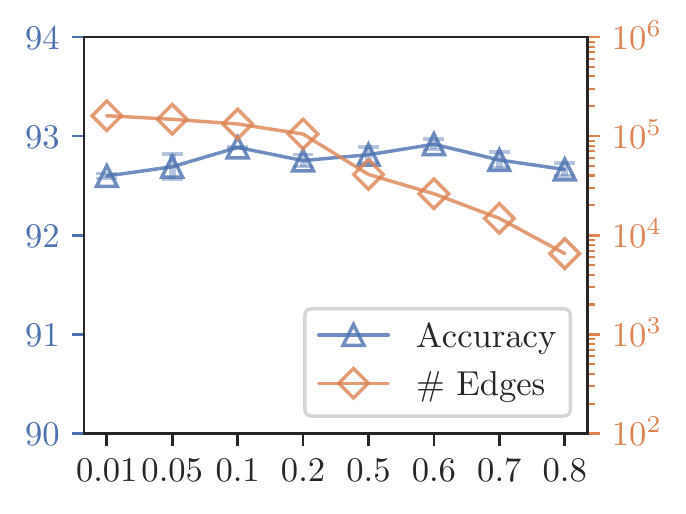}
		\label{fig:sensitivity-nd}
	}\qquad
	\subfloat[Edge removing probability]{
		\includegraphics[scale=0.53]{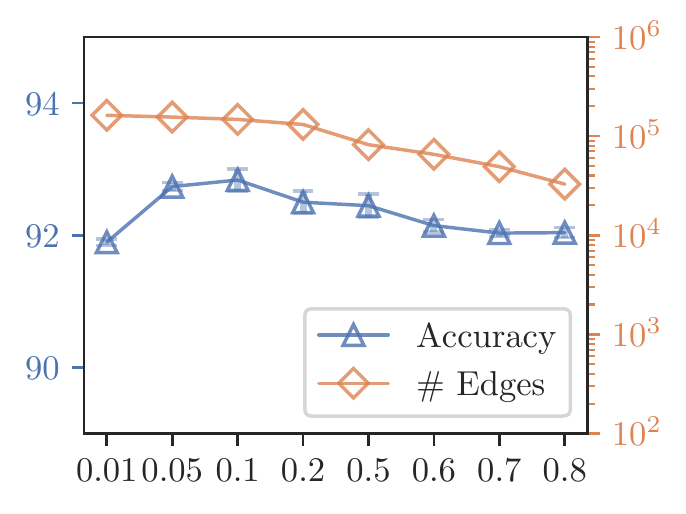}
		\label{fig:sensitivity-er}
	}\qquad
	\subfloat[Edge adding probability]{
		\includegraphics[scale=0.53]{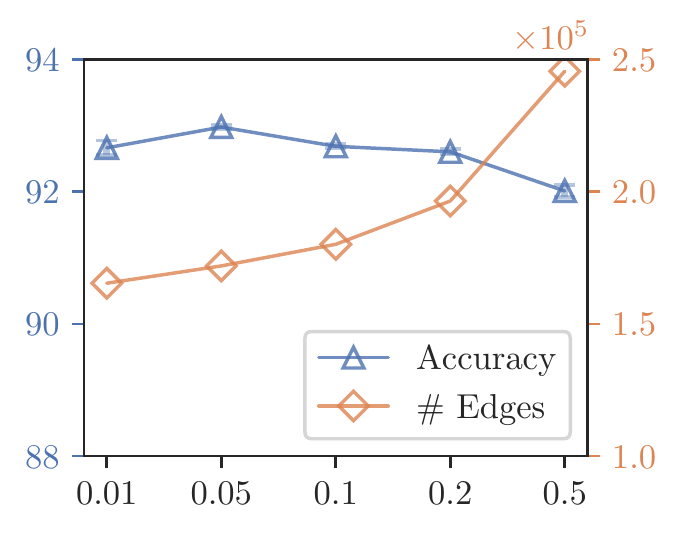}
		\label{fig:sensitivity-ea}
	}
	\caption{Sensitivity analysis with various topology augmentation probabilities on the CS dataset.}
	\label{fig:sensitivity}
	\vskip -1em
\end{figure}

To see how sparsity of the resulting views affect the performance, we further conduct sensitivity analysis on three models with ND, ER, and EA augmentations respectively by varying the dropping/adding probabilities on the CS dataset. The prediction accuracy along with the total number of edges in the produced graphs is plotted in Figure \ref{fig:sensitivity}.
From Figure \ref{fig:sensitivity-nd} and \ref{fig:sensitivity-er}, we observe that model performance improves as more nodes or edges are dropped and degenerates when the removal probability is overly high. As seen in Figure \ref{fig:sensitivity-ea}, the performance of EA augmentation downgrades greatly when more edges are added.
In general, these results accord with our observations that many real-world graphs are inherently sparse \cite{Jin:2020br,Zhu:2021ue}. When too many edges are added, they connect nodes that are semantically unrelated, bringing noise to the generated graph view and thus deteriorating the quality of learned embeddings.

\textbf{Observation 2. Feature augmentations bring additional benefits to GCL. Compositional augmentations at both structure and attribute level benefit GCL most.}

\begin{figure}
	\includegraphics[width=\linewidth]{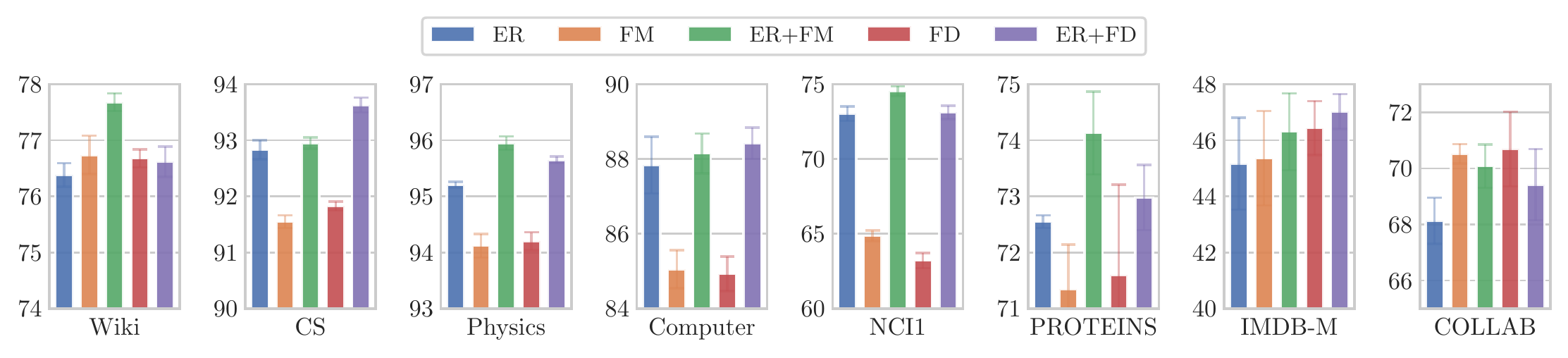}\\
	\includegraphics[width=\linewidth]{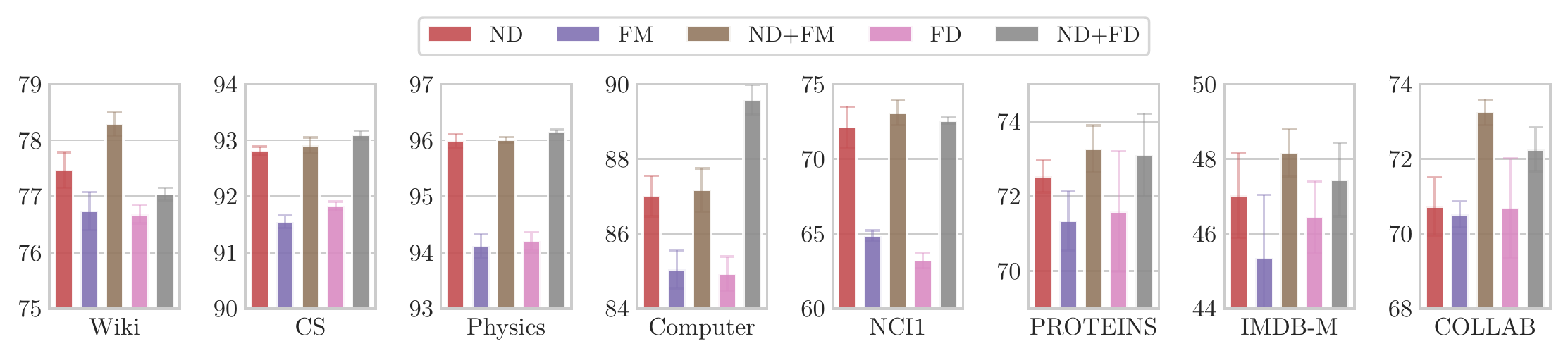}
	\caption{Compositional augmentations using both topology and feature augmentations.}
	\label{fig:structure-feature-augmentations}
	\vskip -0.5em
\end{figure}

From Table \ref{tab:topology-feature-augmentations}, we observe that the performance of models employing feature augmentations solely is inferior to that use topology augmentations but still higher than the baseline performance.
The improvements brought by the two feature augmentation schemes could be explained from the fact that FM and FD resemble applying the dropout \cite{Srivastava:2014cg} technique on the input layer. Also, we see that in general FM slightly outperforms FD, which suggests the use of a shared feature mask for all node features, though their performance difference is not that significant.
Furthermore, Figure \ref{fig:structure-feature-augmentations} presents the results in the case of hybrid augmentations at both topology and feature levels. It is clear that the use of feature augmentation in addition to structure augmentation benefits GCL, demonstrating that both topology and structures are important for learning graph representations.

\textbf{Observation 3. Deterministic augmentation schemes should be accompanied by stochastic augmentations.}

\begin{figure}
	\includegraphics[width=\linewidth]{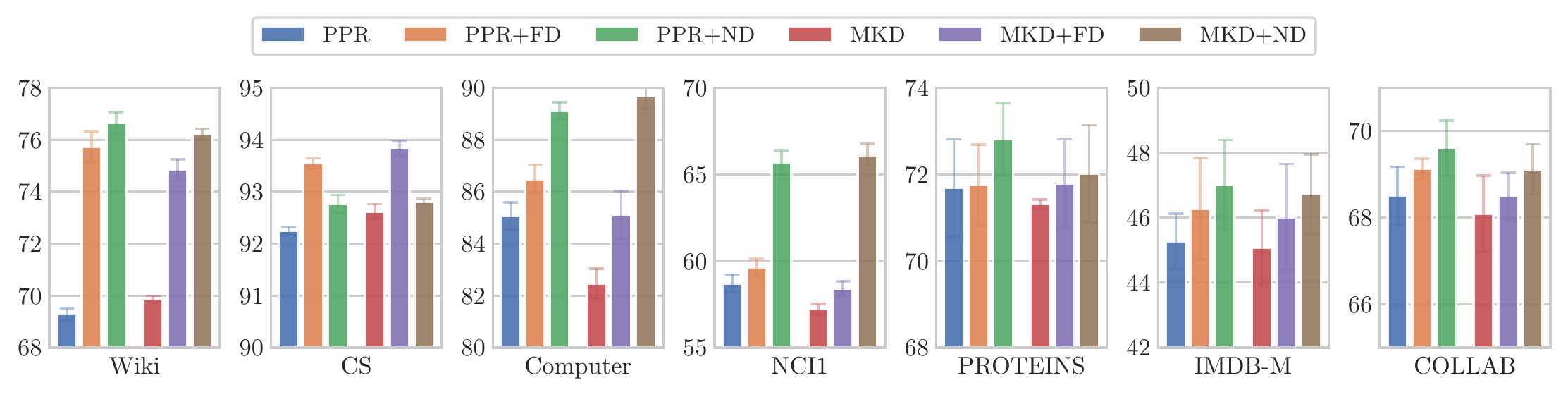}
	\caption{Composition of stochastic and deterministic augmentations.}
	\label{fig:deterministic-stochastic-augmentations}
	\vskip -1.5em
\end{figure}

In Table \ref{tab:topology-feature-augmentations}, we find that solely using two deterministic augmentation functions PPR and MDK does not always result in promising performance. We also find that prior studies \cite{Hassani:2020un} usually leverage a stochastic augmentation after deterministic augmentations. Therefore, we utilize another set of joint augmentations of stochastic and deterministic augmentations, where the performance is summarized in Figure \ref{fig:deterministic-stochastic-augmentations}.
It is seen that the joint scheme improves the vanilla deterministic augmentations by a large margin.
Recall that our contrastive objective is essentially aimed to discriminate between samples from the data distribution and samples from some noise distributions \cite{Gutmann:2012eq,Mnih:2013to}. Therefore, a stochastic augmentation scheme is needed to better approximate that noise distribution.

\subsection{Contrasting Modes and Contrastive Objectives}
\label{sec:contrasting-modes-and-objectives}

\begin{tcolorbox}
\resizebox{\linewidth}{!}{
Augmentations: ND + FM \quad Contrasting mode: \textit{eval} \quad Objectives: \textit{eval} \quad Negative mining strategy: None
}

\end{tcolorbox}
The next experiments are concerned with how contrasting modes and contrastive objectives affect the model performance.
We train the model with different contrasting modes and contrastive objectives, with the topology augmentation set to ND and the feature augmentation to FM. Except that embedding sizes are fixed, other hyperparameters are tuned to obtain the best performance under each experiment for fair comparison.

We first experiment with negative-sample-based contrastive objectives: Information Noice Contrastive Estimation (InfoNCE), Jensen-Shannon Divergence (JSD), and Triplet Margin loss (TM) losses.
Besides contrastive objectives that rely on negative samples, we also investigate three negative-sample-free objectives: Bootstrapping Latent (BL) loss, Barlow Twins (BT) loss, and VICReg loss.
Table \ref{tab:contrastive-objective-and-mode} presents the experimental results on unsupervised classification tasks.
It should be noted that Barlow Twins (BT) and VICReg losses need to compute the covariance of latent embedding vectors and thus these two objectives are not compatible with the global-local mode.

\begin{table}
	\centering
	\caption{Performance with different contrastive objectives and contrastive modes. L--L, G--L, and G--G denote local-local, global-local, and global-global contrasting modes. The best performing results for objectives (row-wise) and contrasting modes (column-wise) are highlighted in boldface and underline respectively.}
	\label{tab:contrastive-objective-and-mode}
	\subfloat[Unsupervised node classification]{
	\resizebox{\linewidth}{!}{
	\begin{tabular}{ccccccccc}
	\toprule
	\multirow{2.5}[0]{*}{Obj.} & \multicolumn{2}{c}{Wiki} & \multicolumn{2}{c}{CS} & \multicolumn{2}{c}{Physics} & \multicolumn{2}{c}{Computer} \\
	\cmidrule(lr){2-3} \cmidrule(lr){4-5} \cmidrule(lr){6-7} \cmidrule(lr){8-9}
	& L--L & G--L & L--L & G--L & L--L & G--L & L--L & G--L \\
	\midrule
	InfoNCE & \cellcolor{gray!20}\underline{\textbf{79.09{\footnotesize ±0.15}}} & 77.73{\footnotesize ±0.94} & \cellcolor{gray!20}\underline{\textbf{92.45{\footnotesize ±0.83}}} & 90.60{\footnotesize ±0.06} & \cellcolor{gray!20}\underline{\textbf{95.95{\footnotesize ±0.92}}} & 93.23{\footnotesize ±0.96} & \cellcolor{gray!20}\underline{\textbf{88.15{\footnotesize ±0.59}}} & 76.24{\footnotesize ±0.93} \\
	JSD   & \textbf{78.83{\footnotesize ±0.95}} & \underline{78.71{\footnotesize ±0.19}} & \textbf{92.18{\footnotesize ±1.00}} & \underline{91.31{\footnotesize ±0.62}} & \textbf{94.32{\footnotesize ±0.28}} & \underline{94.12{\footnotesize ±0.04}} & \textbf{82.02{\footnotesize ±0.76}} & \underline{78.27{\footnotesize ±0.05}} \\
	TM    & \textbf{78.42{\footnotesize ±0.88}} & 76.53{\footnotesize ±0.85} & \textbf{91.91{\footnotesize ±0.31}} & 90.11{\footnotesize ±0.61} & \textbf{94.11{\footnotesize ±0.60}} & 92.78{\footnotesize ±0.12} & 69.67{\footnotesize ±0.88} & \textbf{76.38{\footnotesize ±0.75}} \\
	\midrule
	BL    & 76.83{\footnotesize ±0.80}  & 75.34{\footnotesize ±0.43}  & 93.10{\footnotesize ±0.94}  & 88.55{\footnotesize ±0.43}  & 94.81{\footnotesize ±0.98}  & 94.09{\footnotesize ±0.83}  & \cellcolor{gray!20}\textbf{87.79{\footnotesize ±0.94}} & 85.43{\footnotesize ±0.23}  \\
    BT & 80.41{\footnotesize ±0.15}  & ---   & \cellcolor{gray!20}\textbf{94.16{\footnotesize ±0.02}} & ---   & \cellcolor{gray!20}\textbf{96.55{\footnotesize ±0.12}} & ---   & 86.86{\footnotesize ±0.97}  & --- \\
    VICReg & \cellcolor{gray!20}\textbf{80.79{\footnotesize ±0.12}} & ---   & 93.46{\footnotesize ±0.08}  & ---   & 95.59{\footnotesize ±0.23}  & ---   & 86.39{\footnotesize ±0.32}  & --- \\
	\bottomrule
	\end{tabular}
	}
	\label{tab:contrastive-objective-and-mode-node}
	}
	\\
	\subfloat[Unsupervised graph classification]{
	\resizebox{\linewidth}{!}{
	\begin{tabular}{ccccccccccccc}
	\toprule
	\multirow{2.5}[0]{*}{Obj.} & \multicolumn{3}{c}{NCI1} & \multicolumn{3}{c}{PROTEINS} & \multicolumn{3}{c}{IMDB-M} & \multicolumn{3}{c}{COLLAB} \\
	\cmidrule(lr){2-4} \cmidrule(lr){5-7} \cmidrule(lr){8-10} \cmidrule(lr){11-13}
	& L--L  & G--L  & G--G  & L--L  & G--L  & G--G & L--L  & G--L  & G--G  & L--L  & G--L  & G--G \\
	\midrule
	InfoNCE & 73.10{\footnotesize ±0.83} & 72.35{\footnotesize ±0.21} & \cellcolor{gray!20}\underline{\textbf{73.95{\footnotesize ±0.89}}} & 73.28{\footnotesize ±0.62} & 71.57{\footnotesize ±0.92} & \cellcolor{gray!20}\underline{\textbf{75.73{\footnotesize ±0.09}}} & 48.16{\footnotesize ±0.64} & 47.36{\footnotesize ±0.48} & \cellcolor{gray!20}\underline{\textbf{49.69{\footnotesize ±0.44}}} & \underline{73.25{\footnotesize ±0.34}} & 70.92{\footnotesize ±0.22} & \cellcolor{gray!20}\underline{\textbf{73.72{\footnotesize ±0.12}}} \\
	JSD & \underline{\textbf{73.56{\footnotesize ±0.32}}} & \underline{73.29{\footnotesize ±0.31}} & 70.93{\footnotesize ±0.17} & \underline{\textbf{73.88{\footnotesize ±0.31}}} & \underline{73.15{\footnotesize ±0.42}} & 73.67{\footnotesize ±0.45} & 48.31{\footnotesize ±1.17} & \underline{\textbf{48.61{\footnotesize ±1.21}}} & 48.31{\footnotesize ±1.35} & 70.40{\footnotesize ±0.31} & \underline{\textbf{72.62{\footnotesize ±0.35}}} & 71.60{\footnotesize ±0.32} \\
	TM & \textbf{72.43{\footnotesize ±0.21}} & 71.21{\footnotesize ±0.19} & 72.31{\footnotesize ±0.22} & 72.17{\footnotesize ±0.51} & 72.13{\footnotesize ±1.48} & \textbf{73.78{\footnotesize ±0.47}} & \underline{48.38{\footnotesize ±0.20}} & 47.75{\footnotesize ±1.24} & \textbf{48.58{\footnotesize ±0.62}} & 68.85{\footnotesize ±0.45} & 69.47{\footnotesize ±0.20} & \textbf{72.97{\footnotesize ±0.47}} \\
	\midrule
	BL    & \cellcolor{gray!20}\underline{\textbf{77.22{\footnotesize ±0.13}}} & 75.97{\footnotesize ±0.23}  & {\textbf{76.70{\footnotesize ±0.31}}} & \textbf{77.75{\footnotesize ±0.43}} & 77.32{\footnotesize ±0.21}  & \cellcolor{gray!20}\underline{\textbf{78.17{\footnotesize ±0.59}}} & {\textbf{54.64{\footnotesize ±0.43}}} & {54.21{\footnotesize ±1.01}} & \cellcolor{gray!20}{\underline{\textbf{55.32{\footnotesize ±0.21}}}} & \textbf{73.95{\footnotesize ±0.25}} & 73.35{\footnotesize ±0.24}  & \cellcolor{gray!20}\underline{\textbf{74.92{\footnotesize ±0.33}}} \\
    BT & \underline{72.49{\footnotesize ±0.22}} & ---   & 70.53{\footnotesize ±1.11}  & \underline{74.87{\footnotesize ±0.68}} & ---           & 74.38{\footnotesize ±0.56}  & 48.50{\footnotesize ±0.65}  & ---           & \underline{49.53{\footnotesize ±0.42}} & 71.70{\footnotesize ±0.53}  & ---   & \underline{73.00{\footnotesize ±0.42}} \\
    VICReg & \underline{72.31{\footnotesize ±0.34}} & ---   & 71.60{\footnotesize ±0.36}  & \underline{74.61{\footnotesize ±1.15}} & ---           & 74.38{\footnotesize ±0.57}  & 46.75{\footnotesize ±1.47}  & ---           & \underline{50.28{\footnotesize ±0.55}} & 68.88{\footnotesize ±0.34}  & ---   & \underline{72.50{\footnotesize ±0.31}} \\
	\bottomrule
	\end{tabular}
	}
	\label{tab:contrastive-objective-and-mode-graph}
	}
	\vskip -3em
\end{table}

\textbf{Observation 4. Same-scale contrasting generally performs better. Downstream tasks of different granularities favor different contrasting modes.}

What stands out from the table is that contrasting local-local pairs achieves the best performance on node-level classification, while the global-global mode performs better on graph-level tasks.
This suggests us to use same-scale contrasting modes, which is consistent with the practices used in recent literature \cite{He:2020tu,Chen:2020wj,You:2020ut,Zhu:2020vf,Zhu:2021wh,Chen:2020uu,Verma:2020vv,Wu:2018kw,Henaff:2020ta}.
A possible explanation for this result is that, in global-local contrasting, all node embeddings within the graph constitute the positive samples for each graph embedding. In other words, the global-local mode pulls every node-graph pair together in the embedding space, which may lead to suboptimal performance in downstream tasks.

Furthermore, the contrasting mode should be chosen according to the granularity of end task, i.e. local-local for node-level tasks and global-global for graph-level tasks.
In accordance with the presented results, recent studies \cite{Xie:2021wr,Cole:2021ug,Kotar:2021wh} have made a similar finding for learning visual representations. They demonstrate that being pretrained on instance-level pretext tasks (i.e. contrasting image-level embeddings in the same batch), current CL models achieve suboptimal performance in fine-grained downstream tasks, e.g., semantic segment that requires pixel-level details.

\textbf{Observation 5. Among negative-sample-based objectives, the use of InfoNCE objective leads to consistent improvements across all settings.}

Table \ref{tab:contrastive-objective-and-mode-node} indicates that InfoNCE achieves the best performance among contrastive objectives that need negative samples, which is demonstrated to be effective by many recent methods \cite{Bachman:2019wp,Grill:2020uc,Chen:2020wj,Tian:2019vw,Tian:2020vw}.
\begin{wrapfigure}{r}{0.28\textwidth}
	\centering
	\includegraphics[width=0.27\textwidth]{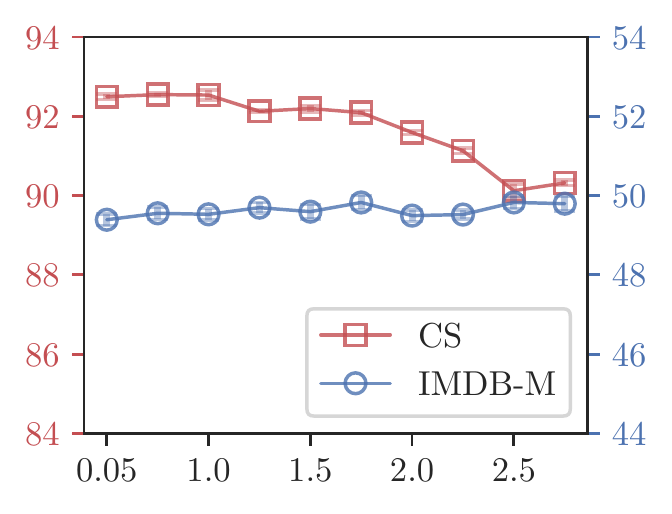}
	\captionsetup{font=scriptsize}
	\caption{Sensitivity analysis of the parameter \(\tau\) in the InfoNCE objective.}
	\label{fig:sensitity-tau}
	\vskip 0.5em
\end{wrapfigure}
Among these negative-sampling-based objectives, recent studies have already revealed that the InfoNCE loss has an intrinsic ability to perform hard negative sampling \cite{Khosla:2020tr}, which may explain its superior performance compared to other objectives.
Particularly, one very recent study in computer vision \cite{Wang:2021vx} shows that the use of a temperature parameter \(\tau\) in the InfoNCE objective acts as an adjustment factor to exert penalties on hard negative samples.
To verify this on GCL, we further conduct sensitivity analysis on this temperature parameter on both node (CS) and graph (IMDB-M) datasets as shown in Figure \ref{fig:sensitity-tau}. We observe that with the increase of \(\tau\), the performance improves at first and downgrades later, with not much performance fluctuation. According to \citet{Wang:2021vx}, the InfoNCE objective pays less attention to hard negatives as \(\tau\) increases. This hardness-aware behavior demonstrates the importance of \emph{striking a balance} between separation of hardest negative samples (\(\tau \rightarrow 0^+\)) and global uniformity (\(\tau \rightarrow \infty\)) on GCL.

\textbf{Observation 6. Bootstrapping Latent and Barlow Twins losses obtain promising performance on par with their negative-sample-based counterparts yet reduce the computational burden without explicit negative samples.}

\begin{wraptable}{r}{0.28\textwidth}
 	\centering
  	\captionsetup{font=scriptsize}
	\caption{Memory usage (MB) on the PROTEINS dataset of different contrastive objectives.}
  	\resizebox{\linewidth}{!}{
    \begin{tabular}{cccc}
    \toprule
    Obj.  & L--L  & G--L  & G--G \\
    \midrule
    InfoNCE & 6,311  & 2,977  & 2,271 \\
    JSD   & 6,309  & 2,845  & 2,269 \\
    TM    & 6,271  & 2,977  & 2,269 \\
    BL    & 2,235  & 2,247  & 2,187 \\
    BT & 2,419 & --- & 2,201 \\
    VICReg & 2,465 & --- & 2,232 \\
    \bottomrule
    \end{tabular}
    }
	\label{tab:memory-usage}
\end{wraptable}
Surprisingly, we find the performance obtained by employing these negative-sample-free objectives sometimes even surpasses their negative-sample-based counterparts, which suggests a promising future direction of more efficient solutions free of negative samples to GCL.
As opposed to negative-sample-based objectives, the BL, BT, and VICReg losses eschew the need of explicit negative samples and thus greatly reduce the computational burden. To see this, we summarize the memory consumption in Table \ref{tab:memory-usage}, from which we clearly observe that these three losses use much less memory than other objectives without negative samples.

\subsection{Negative Mining Strategies}
\label{sec:negative-mining-strategies}

\begin{tcolorbox}
\resizebox{\linewidth}{!}{
Augmentations: ND + FM\quad{}Contrasting mode: InfoNCE\quad{} Objectives: L--L\quad{}Negative mining strategy: \textit{eval}
}
\end{tcolorbox}

We firstly probe the explicit use of negative mining strategies on top of contrastive objectives, which essentially measure similarity (hardness) of each negative pair and upweights hard negative samples. Among the examined negative mining strategies, DCL develops debasing terms to select truly negative samples so as to avoid contrasting same-label instances; other strategies propose to upweight hard negative samples (points that are difficult to distinguish from an anchor) and remove easy ones that are less informative to improve the discriminative power of the GCL model.

\textbf{Observation 7. Existing negative mining techniques based on calculating embedding similarities bring limited benefit to GCL.}

We train four models on three node classification datasets (due to out-of-memory error on Physics) using the local-local mode with the InfoNCE objective.
The performance comparison of different
\begin{wrapfigure}{r}{0.36\textwidth}
	\centering
	\includegraphics[width=0.35\textwidth]{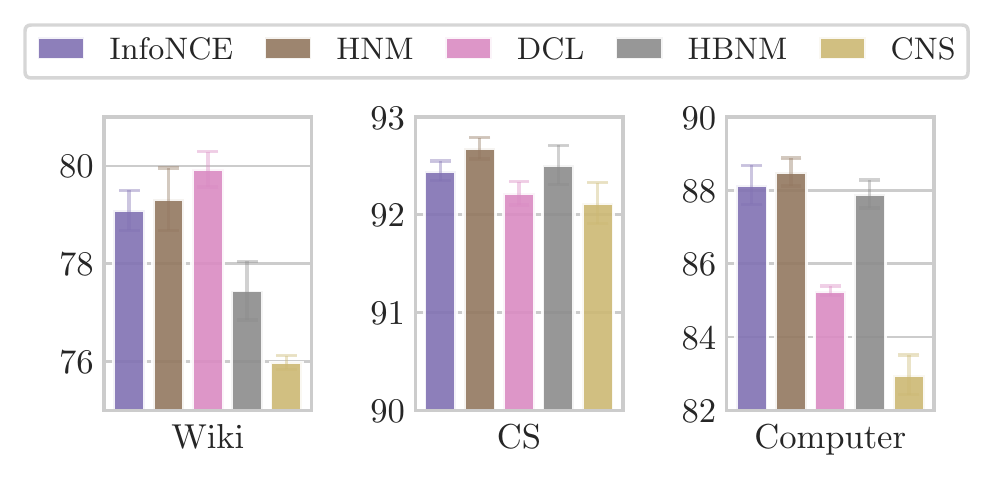}
	\vskip -5pt
	\captionsetup{font=scriptsize}
	\caption{Performance of negative mining strategies.}
	\label{fig:negative-mining}
\end{wrapfigure}
negative mining strategies is shown in Figure \ref{fig:negative-mining}. 
It is apparent that the four negative mining strategies studied in this experiment bring limited improvements to GCL. Although slight improvements can be observed under certain hyperparameter configurations, which proves the explicit modeling of hard negatives is useful for GCL, their performance in general rarely matches the best performance reported with the other contrastive objectives (cf. Table \ref{tab:contrastive-objective-and-mode-node}).

We note that in existing formulation of negative sampling techniques, the sample hardness is measured by inner product of sample embeddings. Since we are in a fully unsupervised setting, no label (i.e. class) information can be accessed during training. Under existing contrasting modes, for one anchor sample, the contrastive objective pushes all different representations away, \emph{irrespective of their real semantic relations}.
What is even worse, %
most GNN models tend to produce similar embeddings for neighboring nodes regardless of semantic classes \cite{Li:2018wc,Chen:2020cn,Zhu:2020wn,Yan:2021wz}, which may further bias the selection of hard negatives.
Therefore, we argue that there is disagreement between semantic similarity and example hardness. By selecting hard negative samples merely according to similarity measure of embeddings, these hard negatives are potentially \emph{positive samples}, which produces adverse learning signals to the contrastive objective.
For an illustrative example and more detailed discussions, please refer to Appendix \ref{sec:discussion-negative-mining}.
We find that our discovery reminiscent to one very recent study in visual CL \cite{Wang:2021vx}, which recommends an adaptive scheduling scheme for the temperature parameter when using InfoNCE as the contrastive objective, such that \emph{hard but false negatives} could be tolerated as the training progresses.

\subsection{Additional Experiments}
We perform additional experiments on large-scale datasets and ablation studies on the batch normalization module used in the bootstrapping latent loss.
Due to space constraints, we briefly discuss our findings as follows. Readers of interest may refer to Appendix \ref{sec:additional-experiments} for detailed discussions.

\textbf{More evaluation tasks and metrics on large-scale datasets with edge features.}
Currently, since the experiments are limited to medium-scale datasets, we conduct a further study assessing how existing GCL work scales to large-scale graphs using two open graph benchmark \cite{Hu:2020wv} datasets.
In the experiments, we examine the performance of different data augmentation due to the presence of edge features, contrasting modes, and contrastive objectives on binary graph classification and graph regression, which are measured in terms of AUC-ROC and MAE, respectively.
It is shown that the experimental results are still in keeping with observations made in the main texts.

\textbf{Ablation studies on the Batch Normalization (BN) for the Bootstrapping Latent (BL) loss.}
Previous work \cite{Richemond:2020ua} hypothesizes that Batch Normalization (BN) compensates for improper initialization for contrastive learning networks, instead of introducing implicit negative samples. To see its impact on GCL performance, we perform ablation studies by using BN or not in three critical components in the BL loss, i.e. the GNN encoder, the projector, and the predictor, on the node classification task.
The results show that using BN in the GNN encoder \emph{solely} is almost sufficient to obtain promising performance.

\section{Conclusion, Limitations, and Outlook}

In this paper, we first present a taxonomy for GCL, where we categorize existing work from four aspects: data augmentations, contrasting modes, contrastive objective, and negative mining strategies.
Then, we analyze the design choices for each GCL component by conducting extensive empirical studies over a comprehensive set of benchmarking tasks and datasets.
Our rigorous empirical study reveal several interesting findings of GCL components that may be helpful for developing future algorithms.
We also provide an open-sourced PyTorch-based library \thesystem to facilitate the implementation of GCL models.
While GCL has already demonstrated strong empirical performance across a variety of downstream tasks, it is still in its infancy with many challenges left widely open.
We hope our work provides several practical guidelines for future research in this vigorous field.

Due to limited space, some limitations of our work need to be acknowledged.
\begin{itemize}[leftmargin=*]
	\item \textbf{Limited design considerations.} In this work, we consider limited design considerations, namely four design dimensions. An issue that is not addressed in this study is what role do many other model-specific factors, e.g., whether to employ a projection head in the InfoNCE objective and what graph encoders should be employed, play in GCL.
	\item \textbf{Limited downstream tasks.} Our empirical study only includes experiments on node- and graph-level classification and graph-level regression; a boarder range of downstream tasks of different granularities, e.g., link prediction and community detection, may be beneficial to draw more convincing conclusions.
	\item \textbf{Lack of theoretical justification.} Our work only presents empirical studies which has thrown up many questions in need of further theoretical justification for better understanding the underlying mechanisms of GCL, for instance, the performance guarantees of certain contrasting modes and how to appropriately measure and select hard negative samples for contrastive objectives in the graph domain.
\end{itemize}

Our empirical findings also suggest several future directions that may be helpful for fully unleashing the power of GCL.
\begin{itemize}[leftmargin=*]
	\item \textbf{Towards automated augmentation.} We understand that topology augmentation is of paramount importance to GCL. However, existing work leverages manually designed ad-hoc augmentation strategies, which may result in suboptimal performance. Recent studies in graph structure learning pave a principled way to learn optimal structures of graph-structured data \cite{Zhu:2021ue}, which we argue could be used for automatically learn augmentation functions suitable for GCL pretext tasks.
	\item \textbf{Understanding the performance gap between pretext and downstream tasks.} We empirically demonstrate the correlation between the choice of end tasks and contrastive objectives, yet calls for a thorough understanding for the performance gap between pretext and downstream tasks. We have found that there is some progress in this regard \cite{Zhu:2020tq}, but it is far from fully explored.
	\item \textbf{Structure-aware negative sampling.} Unlike in computer vision fields, similar visual features may naturally correlate to closer semantic categories, measuring the hardness through embedding similarities in graph-structured data is more difficult. A series of earlier work in network embedding proposes solutions from structural aspects \cite{Wang:2018tw,Ying:2018hq,Yang:2020tm}. However, how to integrate rich structure information for modeling better negative distributions for GCL is still left unexplored.
\end{itemize}

\begin{ack}
The authors would like to thank Yuning You and Liang Zeng for their fruitful discussions on the manuscript.
The authors would also like to thank anonymous reviewers for their helpful feedback.
This work is supported by National Natural Science Foundation of China (61772528, U19B2038).
\end{ack}

\bibliographystyle{unsrtnat}
\bibliography{reference}

\begin{thebibliography}{99}
\providecommand{\natexlab}[1]{#1}
\providecommand{\url}[1]{\texttt{#1}}
\expandafter\ifx\csname urlstyle\endcsname\relax
  \providecommand{\doi}[1]{doi: #1}\else
  \providecommand{\doi}{doi: \begingroup \urlstyle{rm}\Url}\fi

\bibitem[Kipf and Welling(2017)]{Kipf:2016tc}
Thomas~N. Kipf and Max Welling.
\newblock {Semi-Supervised Classification with Graph Convolutional Networks}.
\newblock In \emph{ICLR}, 2017.

\bibitem[Veli{\v c}kovi{\'c} et~al.(2018)Veli{\v c}kovi{\'c}, Cucurull,
  Casanova, Romero, Li{\`o}, and Bengio]{Velickovic:2018we}
Petar Veli{\v c}kovi{\'c}, Guillem Cucurull, Arantxa Casanova, Adriana Romero,
  Pietro Li{\`o}, and Yoshua Bengio.
\newblock {Graph Attention Networks}.
\newblock In \emph{ICLR}, 2018.

\bibitem[Hjelm et~al.(2019)Hjelm, Fedorov, Lavoie-Marchildon, Grewal, Bachman,
  Trischler, and Bengio]{Hjelm:2019uk}
R.~Devon Hjelm, Alex Fedorov, Samuel Lavoie-Marchildon, Karan Grewal, Philip
  Bachman, Adam Trischler, and Yoshua Bengio.
\newblock {Learning Deep Representations by Mutual Information Estimation and
  Maximization}.
\newblock In \emph{ICLR}, 2019.

\bibitem[Bachman et~al.(2019)Bachman, Hjelm, and Buchwalter]{Bachman:2019wp}
Philip Bachman, R.~Devon Hjelm, and William Buchwalter.
\newblock {Learning Representations by Maximizing Mutual Information Across
  Views}.
\newblock In \emph{NeurIPS}, pages 15509--15519, 2019.

\bibitem[He et~al.(2020)He, Fan, Wu, Xie, and Girshick]{He:2020tu}
Kaiming He, Haoqi Fan, Yuxin Wu, Saining Xie, and Ross Girshick.
\newblock {Momentum Contrast for Unsupervised Visual Representation Learning}.
\newblock In \emph{{CVPR}}, pages 9726--9735, 2020.

\bibitem[Caron et~al.(2020)Caron, Misra, Mairal, Goyal, Bojanowski, and
  Joulin]{Caron:2020uv}
Mathilde Caron, Ishan Misra, Julien Mairal, Priya Goyal, Piotr Bojanowski, and
  Armand Joulin.
\newblock {Unsupervised Learning of Visual Features by Contrasting Cluster
  Assignments}.
\newblock In \emph{NeurIPS}, pages 9912--9924, 2020.

\bibitem[Grill et~al.(2020)Grill, Strub, Altch{\'e}, Tallec, Richemond,
  Buchatskaya, Doersch, Pires, Guo, Azar, Piot, Kavukcuoglu, Munos, and
  Valko]{Grill:2020uc}
Jean-Bastien Grill, Florian Strub, Florent Altch{\'e}, Corentin Tallec,
  Pierre~H. Richemond, Elena Buchatskaya, Carl Doersch, Bernardo~Avila Pires,
  Zhaohan~Daniel Guo, Mohammad~Gheshlaghi Azar, Bilal Piot, Koray Kavukcuoglu,
  R{\'e}mi Munos, and Michal Valko.
\newblock {Bootstrap Your Own Latent: A New Approach to Self-Supervised
  Learning}.
\newblock In \emph{NeurIPS}, pages 21271--21284, 2020.

\bibitem[Chen et~al.(2020{\natexlab{a}})Chen, Kornblith, Norouzi, and
  Hinton]{Chen:2020wj}
Ting Chen, Simon Kornblith, Mohammad Norouzi, and Geoffrey~E. Hinton.
\newblock {A Simple Framework for Contrastive Learning of Visual
  Representations}.
\newblock In \emph{ICML}, pages 1597--1607, 2020{\natexlab{a}}.

\bibitem[Qian et~al.(2021)Qian, Meng, Gong, Yang, Wang, Belongie, and
  Cui]{Qian:2021tp}
Rui Qian, Tianjian Meng, Boqing Gong, Ming-Hsuan Yang, Huisheng Wang, Serge~J.
  Belongie, and Yin Cui.
\newblock {Spatiotemporal Contrastive Video Representation Learning}.
\newblock In \emph{CVPR}, 2021.

\bibitem[Klein and Nabi(2020)]{Klein:2020jb}
Tassilo Klein and Moin Nabi.
\newblock {Contrastive Self-Supervised Learning for Commonsense Reasoning}.
\newblock In \emph{ACL}, pages 7517--7523, 2020.

\bibitem[Fang et~al.(2020)Fang, Wang, Zhou, Ding, and Xie]{Fang:2020ux}
Hongchao Fang, Sicheng Wang, Meng Zhou, Jiayuan Ding, and Pengtao Xie.
\newblock {CERT: Contrastive Self-supervised Learning for Language
  Understanding}.
\newblock \emph{arXiv.org}, 2020.

\bibitem[Gao et~al.(2021)Gao, Yao, and Chen]{Gao:2021wf}
Tianyu Gao, Xingcheng Yao, and Danqi Chen.
\newblock {SimCSE: Simple Contrastive Learning of Sentence Embeddings}.
\newblock \emph{arXiv.org}, April 2021.

\bibitem[Veli{\v c}kovi{\'c} et~al.(2019)Veli{\v c}kovi{\'c}, Fedus, Hamilton,
  Li{\`o}, Bengio, and Hjelm]{Velickovic:2019tu}
Petar Veli{\v c}kovi{\'c}, William Fedus, William~L. Hamilton, Pietro Li{\`o},
  Yoshua Bengio, and R.~Devon Hjelm.
\newblock {Deep Graph Infomax}.
\newblock In \emph{ICLR}, 2019.

\bibitem[Hassani and Khasahmadi(2020)]{Hassani:2020un}
Kaveh Hassani and Amir~Hosein Khasahmadi.
\newblock {Contrastive Multi-View Representation Learning on Graphs}.
\newblock In \emph{ICML}, pages 3451--3461, 2020.

\bibitem[Peng et~al.(2020)Peng, Huang, Luo, Zheng, Rong, Xu, and
  Huang]{Peng:2020gw}
Zhen Peng, Wenbing Huang, Minnan Luo, Qinghua Zheng, Yu~Rong, Tingyang Xu, and
  Junzhou Huang.
\newblock {Graph Representation Learning via Graphical Mutual Information
  Maximization}.
\newblock In \emph{WWW}, pages 259--270, 2020.

\bibitem[You et~al.(2020)You, Chen, Sui, Chen, Wang, and Shen]{You:2020ut}
Yuning You, Tianlong Chen, Yongduo Sui, Ting Chen, Zhangyang Wang, and Yang
  Shen.
\newblock {Graph Contrastive Learning with Augmentations}.
\newblock In \emph{NeurIPS}, pages 5812--5823, 2020.

\bibitem[Zhu et~al.(2020{\natexlab{a}})Zhu, Xu, Yu, Liu, Wu, and
  Wang]{Zhu:2020vf}
Yanqiao Zhu, Yichen Xu, Feng Yu, Qiang Liu, Shu Wu, and Liang Wang.
\newblock {Deep Graph Contrastive Representation Learning}.
\newblock In \emph{GRL+@ICML}, 2020{\natexlab{a}}.

\bibitem[Zhu et~al.(2021{\natexlab{a}})Zhu, Xu, Yu, Liu, Wu, and
  Wang]{Zhu:2021wh}
Yanqiao Zhu, Yichen Xu, Feng Yu, Qiang Liu, Shu Wu, and Liang Wang.
\newblock {Graph Contrastive Learning with Adaptive Augmentation}.
\newblock In \emph{WWW}, pages 2069--2080, 2021{\natexlab{a}}.

\bibitem[Sun et~al.(2020)Sun, Hoffmann, Verma, and Tang]{Sun:2020vi}
Fan-Yun Sun, Jordan Hoffmann, Vikas Verma, and Jian Tang.
\newblock {InfoGraph: Unsupervised and Semi-supervised Graph-Level
  Representation Learning via Mutual Information Maximization}.
\newblock In \emph{ICLR}, 2020.

\bibitem[Xie et~al.(2021{\natexlab{a}})Xie, Xu, Wang, and Ji]{Xie:2021uv}
Yaochen Xie, Zhao Xu, Zhengyang Wang, and Shuiwang Ji.
\newblock {Self-Supervised Learning of Graph Neural Networks: A Unified
  Review}.
\newblock \emph{arXiv.org}, February 2021{\natexlab{a}}.

\bibitem[Liu et~al.(2021)Liu, Pan, Jin, Zhou, Xia, and Yu]{Liu:2021wc}
Yixin Liu, Shirui Pan, Ming Jin, Chuan Zhou, Feng Xia, and Philip~S. Yu.
\newblock {Graph Self-Supervised Learning: A Survey}.
\newblock \emph{arXiv.org}, February 2021.

\bibitem[Wu et~al.(2021{\natexlab{a}})Wu, Lin, Gao, Tan, and Li]{Wu:2021wg}
Lirong Wu, Haitao Lin, Zhangyang Gao, Cheng Tan, and Stan~Z. Li.
\newblock {Self-supervised on Graphs: Contrastive, Generative, or Predictive}.
\newblock \emph{arXiv.org}, May 2021{\natexlab{a}}.

\bibitem[Khosla et~al.(2020)Khosla, Teterwak, Wang, Sarna, Tian, Isola,
  Maschinot, Liu, and Krishnan]{Khosla:2020tr}
Prannay Khosla, Piotr Teterwak, Chen Wang, Aaron Sarna, Yonglong Tian, Phillip
  Isola, Aaron Maschinot, Ce~Liu, and Dilip Krishnan.
\newblock {Supervised Contrastive Learning}.
\newblock In \emph{NeurIPS}, pages 18661--18673, 2020.

\bibitem[Qiu et~al.(2020)Qiu, Chen, Dong, Zhang, Yang, Ding, Wang, and
  Tang]{Qiu:2020gq}
Jiezhong Qiu, Qibin Chen, Yuxiao Dong, Jing Zhang, Hongxia Yang, Ming Ding,
  Kuansan Wang, and Jie Tang.
\newblock {GCC: Graph Contrastive Coding for Graph Neural Network
  Pre-Training}.
\newblock In \emph{KDD}, pages 1150--1160, 2020.

\bibitem[Thakoor et~al.(2021)Thakoor, Tallec, Azar, Munos, Veli{\v c}kovi{\'c},
  and Valko]{Thakoor:2021tl}
Shantanu Thakoor, Corentin Tallec, Mohammad~Gheshlaghi Azar, R{\'e}mi Munos,
  Petar Veli{\v c}kovi{\'c}, and Michal Valko.
\newblock {Bootstrapped Representation Learning on Graphs}.
\newblock \emph{arXiv.org}, February 2021.

\bibitem[Bielak et~al.(2021)Bielak, Kajdanowicz, and Chawla]{Bielak:2021uv}
Piotr Bielak, Tomasz Kajdanowicz, and Nitesh~V. Chawla.
\newblock {Graph Barlow Twins: A Self-Supervised Representation Learning
  Framework for Graphs}.
\newblock \emph{arXiv.org}, June 2021.

\bibitem[Zhang et~al.(2020)Zhang, Hu, Subramonian, and Sun]{Zhang:2020ts}
Shichang Zhang, Ziniu Hu, Arjun Subramonian, and Yizhou Sun.
\newblock {Motif-Driven Contrastive Learning of Graph Representations}.
\newblock \emph{arXiv.org}, December 2020.

\bibitem[Mavromatis and Karypis(2021)]{Mavromatis:2021jr}
Costas Mavromatis and George Karypis.
\newblock {Graph InfoClust: Maximizing Coarse-Grain Mutual Information in
  Graphs}.
\newblock In \emph{PAKDD}, pages 541--553, 2021.

\bibitem[Shchur et~al.(2018)Shchur, Mumme, Bojchevski, and
  G{\"u}nnemann]{Shchur:2018vv}
Oleksandr Shchur, Maximilian Mumme, Aleksandar Bojchevski, and Stephan
  G{\"u}nnemann.
\newblock {Pitfalls of Graph Neural Network Evaluation}.
\newblock In \emph{R2L@NeurIPS}, 2018.

\bibitem[Morris et~al.(2020)Morris, Kriege, Bause, Kersting, Mutzel, and
  Neumann]{Morris:2020wd}
Christopher Morris, Nils~M. Kriege, Franka Bause, Kristian Kersting, Petra
  Mutzel, and Marion Neumann.
\newblock {TUDataset: a Collection of Benchmark Datasets for Learning with
  Graphs}.
\newblock In \emph{GRL+@ICML}, 2020.

\bibitem[Mernyei and Cangea(2020)]{Mernyei:2020wh}
P{\'e}ter Mernyei and Catalina Cangea.
\newblock {Wiki-CS: A Wikipedia-Based Benchmark for Graph Neural Networks}.
\newblock In \emph{GRL+@ICML}, July 2020.

\bibitem[Wale and Karypis(2006)]{Wale:2006kv}
Nikil Wale and George Karypis.
\newblock {Comparison of Descriptor Spaces for Chemical Compound Retrieval and
  Classification}.
\newblock In \emph{ICDM}, pages 678--689, 2006.

\bibitem[Borgwardt et~al.(2005)Borgwardt, Ong, Sch{\"o}nauer, Vishwanathan,
  Smola, and Kriegel]{Borgwardt:2005jk}
Karsten~M. Borgwardt, Cheng~Soon Ong, Stefan Sch{\"o}nauer, S.~V.~N.
  Vishwanathan, Alexander~J. Smola, and Hans-Peter Kriegel.
\newblock {Protein Function Prediction via Graph Kernels}.
\newblock In \emph{ISMB}, pages 47--56, 2005.

\bibitem[Yanardag and Vishwanathan(2015)]{Yanardag:2015fm}
Pinar Yanardag and S.~V.~N. Vishwanathan.
\newblock {Deep Graph Kernels}.
\newblock In \emph{KDD}, pages 1365--1374, 2015.

\bibitem[Xu et~al.(2019)Xu, Hu, Leskovec, and Jegelka]{Xu:2019ty}
Keyulu Xu, Weihua Hu, Jure Leskovec, and Stefanie Jegelka.
\newblock {How Powerful are Graph Neural Networks?}
\newblock In \emph{ICLR}, 2019.

\bibitem[Jin et~al.(2020)Jin, Ma, Liu, Tang, Wang, and Tang]{Jin:2020br}
Wei Jin, Yao Ma, Xiaorui Liu, Xianfeng Tang, Suhang Wang, and Jiliang Tang.
\newblock {Graph Structure Learning for Robust Graph Neural Networks}.
\newblock In \emph{KDD}, pages 66--74. ACM, 2020.

\bibitem[Zhu et~al.(2021{\natexlab{b}})Zhu, Xu, Zhang, Liu, Wu, and
  Wang]{Zhu:2021ue}
Yanqiao Zhu, Weizhi Xu, Jinghao Zhang, Qiang Liu, Shu Wu, and Liang Wang.
\newblock {Deep Graph Structure Learning for Robust Representations: A Survey}.
\newblock \emph{arXiv.org}, March 2021{\natexlab{b}}.

\bibitem[Srivastava et~al.(2014)Srivastava, Hinton, Krizhevsky, Sutskever, and
  Salakhutdinov]{Srivastava:2014cg}
Nitish Srivastava, Geoffrey~E. Hinton, Alex Krizhevsky, Ilya Sutskever, and
  Ruslan~R. Salakhutdinov.
\newblock {Dropout: A Simple Way to Prevent Neural Networks From Overfitting}.
\newblock \emph{J. Mach. Learn. Res.}, 15\penalty0 (1):\penalty0 1929--1958,
  2014.

\bibitem[Gutmann and Hyv{\"a}rinen(2012)]{Gutmann:2012eq}
Michael Gutmann and Aapo Hyv{\"a}rinen.
\newblock {Noise-Contrastive Estimation of Unnormalized Statistical Models,
  with Applications to Natural Image Statistics}.
\newblock \emph{J. Mach. Learn. Res.}, 13:\penalty0 307--361, 2012.

\bibitem[Mnih and Kavukcuoglu(2013)]{Mnih:2013to}
Andriy Mnih and Koray Kavukcuoglu.
\newblock {Learning Word Embeddings Efficiently with Noise-Contrastive
  Estimation}.
\newblock In \emph{NIPS}, pages 2265--2273, 2013.

\bibitem[Chen and He(2020)]{Chen:2020uu}
Xinlei Chen and Kaiming He.
\newblock {Exploring Simple Siamese Representation Learning}.
\newblock \emph{arXiv.org}, November 2020.

\bibitem[Verma et~al.(2020)Verma, Luong, Kawaguchi, Pham, and Le]{Verma:2020vv}
Vikas Verma, Minh-Thang Luong, Kenji Kawaguchi, Hieu Pham, and Quoc~V. Le.
\newblock {Towards Domain-Agnostic Contrastive Learning}.
\newblock \emph{arXiv.org}, November 2020.

\bibitem[Wu et~al.(2018)Wu, Xiong, Yu, and Lin]{Wu:2018kw}
Zhirong Wu, Yuanjun Xiong, Stella~X. Yu, and Dahua Lin.
\newblock {Unsupervised Feature Learning via Non-Parametric Instance
  Discrimination}.
\newblock In \emph{CVPR}, pages 3733--3742, 2018.

\bibitem[H{\'e}naff et~al.(2020)H{\'e}naff, Srinivas, De~Fauw, Razavi, Doersch,
  Eslami, and van~den Oord]{Henaff:2020ta}
Olivier~J. H{\'e}naff, Aravind Srinivas, Jeffrey De~Fauw, Ali Razavi, Carl
  Doersch, S.~M.~Ali Eslami, and A{\"a}ron van~den Oord.
\newblock {Data-Efficient Image Recognition with Contrastive Predictive
  Coding}.
\newblock In \emph{ICML}, pages 4182--4192, 2020.

\bibitem[Xie et~al.(2021{\natexlab{b}})Xie, Lin, Zhang, Cao, Lin, and
  Hu]{Xie:2021wr}
Zhenda Xie, Yutong Lin, Zheng Zhang, Yue Cao, Stephen Lin, and Han Hu.
\newblock {Propagate Yourself: Exploring Pixel-Level Consistency for
  Unsupervised Visual Representation Learning}.
\newblock In \emph{CVPR}, 2021{\natexlab{b}}.

\bibitem[Cole et~al.(2021)Cole, Yang, Wilber, Mac~Aodha, and
  Belongie]{Cole:2021ug}
Elijah Cole, Xuan Yang, Kimberly Wilber, Oisin Mac~Aodha, and Serge Belongie.
\newblock {When Does Contrastive Visual Representation Learning Work?}
\newblock \emph{arXiv.org}, May 2021.

\bibitem[Kotar et~al.(2021)Kotar, Ilharco, Schmidt, Ehsani, and
  Mottaghi]{Kotar:2021wh}
Klemen Kotar, Gabriel Ilharco, Ludwig Schmidt, Kiana Ehsani, and Roozbeh
  Mottaghi.
\newblock {Contrasting Contrastive Self-Supervised Representation Learning
  Pipelines}.
\newblock In \emph{ICCV}, 2021.

\bibitem[Tian et~al.(2020{\natexlab{a}})Tian, Krishnan, and Isola]{Tian:2019vw}
Yonglong Tian, Dilip Krishnan, and Phillip Isola.
\newblock {Contrastive Multiview Coding}.
\newblock In \emph{ECCV}, pages 776--794, 2020{\natexlab{a}}.

\bibitem[Tian et~al.(2020{\natexlab{b}})Tian, Sun, Poole, Krishnan, Schmid, and
  Isola]{Tian:2020vw}
Yonglong Tian, Chen Sun, Ben Poole, Dilip Krishnan, Cordelia Schmid, and
  Phillip Isola.
\newblock {What Makes for Good Views for Contrastive Learning?}
\newblock In \emph{NeurIPS}, pages 6827--6839, 2020{\natexlab{b}}.

\bibitem[Wang and Liu(2021)]{Wang:2021vx}
Feng Wang and Huaping Liu.
\newblock {Understanding the Behaviour of Contrastive Loss}.
\newblock In \emph{CVPR}, 2021.

\bibitem[Li et~al.(2018)Li, Han, and Wu]{Li:2018wc}
Qimai Li, Zhichao Han, and Xiao-Ming Wu.
\newblock {Deeper Insights into Graph Convolutional Networks for
  Semi-Supervised Learning}.
\newblock In \emph{AAAI}, pages 3538--3545, 2018.

\bibitem[Chen et~al.(2020{\natexlab{b}})Chen, Lin, Li, Li, Zhou, and
  Sun]{Chen:2020cn}
Deli Chen, Yankai Lin, Wei Li, Peng Li, Jie Zhou, and Xu~Sun.
\newblock {Measuring and Relieving the Over-Smoothing Problem for Graph Neural
  Networks from the Topological View}.
\newblock In \emph{AAAI}, pages 3438--3445, 2020{\natexlab{b}}.

\bibitem[Zhu et~al.(2020{\natexlab{b}})Zhu, Xu, Liu, and Wu]{Zhu:2020wn}
Yanqiao Zhu, Weizhi Xu, Qiang Liu, and Shu Wu.
\newblock {When Contrastive Learning Meets Active Learning: A Novel Graph
  Active Learning Paradigm with Self-Supervision}.
\newblock \emph{arXiv.org}, October 2020{\natexlab{b}}.

\bibitem[Yan et~al.(2021)Yan, Hashemi, Swersky, Yang, and Koutra]{Yan:2021wz}
Yujun Yan, Milad Hashemi, Kevin Swersky, Yaoqing Yang, and Danai Koutra.
\newblock {Two Sides of the Same Coin: Heterophily and Oversmoothing in Graph
  Convolutional Neural Networks}.
\newblock \emph{arXiv.org}, February 2021.

\bibitem[Hu et~al.(2020{\natexlab{a}})Hu, Fey, Zitnik, Dong, Ren, Liu, Catasta,
  and Leskovec]{Hu:2020wv}
Weihua Hu, Matthias Fey, Marinka Zitnik, Yuxiao Dong, Hongyu Ren, Bowen Liu,
  Michele Catasta, and Jure Leskovec.
\newblock {Open Graph Benchmark: Datasets for Machine Learning on Graphs}.
\newblock In \emph{NeurIPS}, pages 22118--22133, 2020{\natexlab{a}}.

\bibitem[Richemond et~al.(2020)Richemond, Grill, Altch{\'e}, Tallec, Strub,
  Brock, Smith, De, Pascanu, Piot, and Valko]{Richemond:2020ua}
Pierre~H. Richemond, Jean-Bastien Grill, Florent Altch{\'e}, Corentin Tallec,
  Florian Strub, Andrew Brock, Samuel Smith, Soham De, Razvan Pascanu, Bilal
  Piot, and Michal Valko.
\newblock {BYOL Works Even Without Batch Statistics}.
\newblock \emph{arXiv.org}, October 2020.

\bibitem[Zhu et~al.(2020{\natexlab{c}})Zhu, Xu, Wang, Zhang, Han, and
  Yang]{Zhu:2020tq}
Qi~Zhu, Yidan Xu, Haonan Wang, Chao Zhang, Jiawei Han, and Carl Yang.
\newblock {Transfer Learning of Graph Neural Networks with Ego-graph
  Information Maximization}.
\newblock \emph{arXiv.org}, September 2020{\natexlab{c}}.

\bibitem[Wang et~al.(2018)Wang, Wang, Wang, Zhao, Zhang, Zhang, Xie, and
  Guo]{Wang:2018tw}
Hongwei Wang, Jia Wang, Jialin Wang, Miao Zhao, Weinan Zhang, Fuzheng Zhang,
  Xing Xie, and Minyi Guo.
\newblock {GraphGAN: Graph Representation Learning with Generative Adversarial
  Nets}.
\newblock In \emph{AAAI}, pages 2508--2515, 2018.

\bibitem[Ying et~al.(2018)Ying, He, Chen, Eksombatchai, Hamilton, and
  Leskovec]{Ying:2018hq}
Rex Ying, Ruining He, Kaifeng Chen, Pong Eksombatchai, William~L. Hamilton, and
  Jure Leskovec.
\newblock {Graph Convolutional Neural Networks for Web-Scale Recommender
  Systems}.
\newblock In \emph{KDD}, pages 974--983. ACM, 2018.

\bibitem[Yang et~al.(2020)Yang, Ding, Zhou, Yang, Zhou, and Tang]{Yang:2020tm}
Zhen Yang, Ming Ding, Chang Zhou, Hongxia Yang, Jingren Zhou, and Jie Tang.
\newblock {Understanding Negative Sampling in Graph Representation Learning}.
\newblock In \emph{KDD}, pages 1666--1676. ACM, 2020.

\bibitem[Paszke et~al.(2019)Paszke, Gross, Massa, Lerer, Bradbury, Chanan,
  Killeen, Lin, Gimelshein, Antiga, Desmaison, Kopf, Yang, DeVito, Raison,
  Tejani, Chilamkurthy, Steiner, Fang, Bai, and Chintala]{Paszke:2019vf}
Adam Paszke, Sam Gross, Francisco Massa, Adam Lerer, James Bradbury, Gregory
  Chanan, Trevor Killeen, Zeming Lin, Natalia Gimelshein, Luca Antiga, Alban
  Desmaison, Andreas Kopf, Edward Yang, Zachary DeVito, Martin Raison, Alykhan
  Tejani, Sasank Chilamkurthy, Benoit Steiner, Lu~Fang, Junjie Bai, and Soumith
  Chintala.
\newblock {PyTorch: An Imperative Style, High-Performance Deep Learning
  Library}.
\newblock In \emph{NeurIPS}, pages 8024--8035, 2019.

\bibitem[Fey and Lenssen(2019)]{Fey:2019wv}
Matthias Fey and Jan~Eric Lenssen.
\newblock {Fast Graph Representation Learning with PyTorch Geometric}.
\newblock In \emph{RLGM@ICLR}, 2019.

\bibitem[Hu et~al.(2021)Hu, Fey, Ren, Nakata, Dong, and Leskovec]{Hu:2021wk}
Weihua Hu, Matthias Fey, Hongyu Ren, Maho Nakata, Yuxiao Dong, and Jure
  Leskovec.
\newblock {OGB-LSC: A Large-Scale Challenge for Machine Learning on Graphs}.
\newblock \emph{arXiv.org}, March 2021.

\bibitem[Hu et~al.(2020{\natexlab{b}})Hu, Liu, Gomes, Zitnik, Liang, Pande, and
  Leskovec]{Hu:2020uz}
Weihua Hu, Bowen Liu, Joseph Gomes, Marinka Zitnik, Percy Liang, Vijay Pande,
  and Jure Leskovec.
\newblock {Strategies for Pre-training Graph Neural Networks}.
\newblock In \emph{ICLR}, 2020{\natexlab{b}}.

\bibitem[Krizhevsky et~al.(2012)Krizhevsky, Sutskever, and
  Hinton]{Krizhevsky:2012wl}
Alex Krizhevsky, Ilya Sutskever, and Geoffrey~E. Hinton.
\newblock {ImageNet Classification with Deep Convolutional Neural Networks}.
\newblock In \emph{NIPS}, 2012.

\bibitem[Szegedy et~al.(2015)Szegedy, Liu, Jia, Sermanet, Reed, Anguelov,
  Erhan, Vanhoucke, and Rabinovich]{Szegedy:2015tb}
Christian Szegedy, Wei Liu, Yangqing Jia, Pierre Sermanet, Scott~E. Reed,
  Dragomir Anguelov, Dumitru Erhan, Vincent Vanhoucke, and Andrew Rabinovich.
\newblock {Going Deeper with Convolutions}.
\newblock In \emph{CVPR}, pages 1--9, 2015.

\bibitem[Larsson et~al.(2017)Larsson, Maire, and Shakhnarovich]{Larsson:2017vt}
Gustav Larsson, Michael Maire, and Gregory Shakhnarovich.
\newblock {Colorization as a Proxy Task for Visual Understanding}.
\newblock In \emph{{CVPR}}, pages 840--849, 2017.

\bibitem[Gidaris et~al.(2018)Gidaris, Singh, and Komodakis]{Gidaris:2018wr}
Spyros Gidaris, Praveer Singh, and Nikos Komodakis.
\newblock {Unsupervised Representation Learning by Predicting Image Rotations}.
\newblock In \emph{Proceedings of the 6th International Conference on Learning
  Representations}, 2018.

\bibitem[Wang et~al.(2020)Wang, Wang, Liang, Cai, Liu, and Hooi]{Wang:2020kt}
Yiwei Wang, Wei Wang, Yuxuan Liang, Yujun Cai, Juncheng Liu, and Bryan Hooi.
\newblock {NodeAug: Semi-Supervised Node Classification with Data
  Augmentation}.
\newblock In \emph{KDD}, pages 207--217, 2020.

\bibitem[Guo et~al.(2021)Guo, Du, and Zhao]{Guo:2021ex}
Xiaojie Guo, Yuanqi Du, and Liang Zhao.
\newblock {Deep Generative Models for Spatial Networks}.
\newblock In \emph{KDD}, pages 505--515, 2021.

\bibitem[Du et~al.(2021)Du, Wang, Guo, Cao, Hu, Jiang, Varala, Angirekula, and
  Zhao]{Du:2021gt}
Yuanqi Du, Shiyu Wang, Xiaojie Guo, Hengning Cao, Shujie Hu, Junji Jiang,
  Aishwarya Varala, Abhinav Angirekula, and Liang Zhao.
\newblock {GraphGT: Machine Learning Datasets for Deep Graph Generation and
  Transformation}.
\newblock In \emph{NeurIPS}, 2021.

\bibitem[Tong et~al.(2006)Tong, Faloutsos, and Pan]{Tong:2006jh}
Hanghang Tong, Christos Faloutsos, and Jia-Yu Pan.
\newblock {Fast Random Walk with Restart and Its Applications}.
\newblock In \emph{ICDM}, pages 613--622, 2006.

\bibitem[Page et~al.(1999)Page, Brin, Motwani, and Winograd]{Page:1999wg}
Lawrence Page, Sergey Brin, Rajeev Motwani, and Terry Winograd.
\newblock {The PageRank Citation Ranking: Bringing Order to the Web}.
\newblock Technical report, Stanford InfoLab, November 1999.

\bibitem[Klicpera et~al.(2019)Klicpera, Wei{\ss}enberger, and
  G{\"u}nnemann]{Klicpera:2019vc}
Johannes Klicpera, Stefan Wei{\ss}enberger, and Stephan G{\"u}nnemann.
\newblock {Diffusion Improves Graph Learning}.
\newblock In \emph{NeurIPS}, pages 13333--13345, 2019.

\bibitem[Fouss et~al.(2012)Fouss, Fran{\c c}oisse, Yen, Pirotte, and
  Saerens]{Fouss:2012bf}
Fran{\c c}ois Fouss, Kevin Fran{\c c}oisse, Luh Yen, Alain Pirotte, and Marco
  Saerens.
\newblock {An Experimental Investigation of Kernels on Graphs for Collaborative
  Recommendation and Semisupervised Classification}.
\newblock \emph{Neural Networks}, 31:\penalty0 53--72, 2012.

\bibitem[Zhu and Koniusz(2021)]{Zhu:2021ta}
Hao Zhu and Piotr Koniusz.
\newblock {Simple Spectral Graph Convolution}.
\newblock In \emph{ICLR}, 2021.

\bibitem[van~den Oord et~al.(2018)van~den Oord, Li, and
  Vinyals]{vandenOord:2018ut}
A{\"a}ron van~den Oord, Yazhe Li, and Oriol Vinyals.
\newblock {Representation Learning with Contrastive Predictive Coding}.
\newblock \emph{arXiv.org}, 2018.

\bibitem[Schroff et~al.(2015)Schroff, Kalenichenko, and
  Philbin]{Schroff:2015wo}
Florian Schroff, Dmitry Kalenichenko, and James Philbin.
\newblock {FaceNet: A Unified Embedding for Face Recognition and Clustering}.
\newblock In \emph{CVPR}, pages 815--823, 2015.

\bibitem[Linsker(1988)]{Linsker:1988ho}
Ralph Linsker.
\newblock {Self-Organization in a Perceptual Network}.
\newblock \emph{IEEE Computer}, 21\penalty0 (3):\penalty0 105--117, 1988.

\bibitem[Gutmann and Hyv{\"a}rinen(2010)]{Hyvarinen:2010ua}
Michael Gutmann and Aapo Hyv{\"a}rinen.
\newblock {Noise-Contrastive Estimation: a New Estimation Principle for
  Unnormalized Statistical Models}.
\newblock In \emph{AISTATS}, pages 297--304, 2010.

\bibitem[Nowozin et~al.(2016)Nowozin, Cseke, and Tomioka]{Nowozin:2016uq}
Sebastian Nowozin, Botond Cseke, and Ryota Tomioka.
\newblock {f-GAN: Training Generative Neural Samplers using Variational
  Divergence Minimization}.
\newblock In \emph{NIPS}, pages 271--279, 2016.

\bibitem[Poole et~al.(2019)Poole, Ozair, van~den Oord, Alemi, and
  Tucker]{Poole:2019vk}
Ben Poole, Sherjil Ozair, A{\"a}ron van~den Oord, Alexander~A. Alemi, and
  George Tucker.
\newblock {On Variational Bounds of Mutual Information}.
\newblock In \emph{ICML}, pages 5171--5180, 2019.

\bibitem[Tschannen et~al.(2020)Tschannen, Djolonga, Rubenstein, Gelly, and
  Lucic]{Tschannen:2020uo}
Michael Tschannen, Josip Djolonga, Paul~K. Rubenstein, Sylvain Gelly, and Mario
  Lucic.
\newblock {On Mutual Information Maximization for Representation Learning}.
\newblock In \emph{ICLR}, 2020.

\bibitem[Wang and Isola(2020)]{Wang:2020wi}
Tongzhou Wang and Phillip Isola.
\newblock {Understanding Contrastive Representation Learning through Alignment
  and Uniformity on the Hypersphere}.
\newblock In \emph{ICML}, pages 9929--9939, 2020.

\bibitem[Tian et~al.(2021)Tian, Chen, and Ganguli]{Tian:2021vl}
Yuandong Tian, Xinlei Chen, and Surya Ganguli.
\newblock {Understanding Self-Supervised Learning Dynamics without Contrastive
  Pairs}.
\newblock In \emph{ICML}, 2021.

\bibitem[Kefato and Girdzijauskas(2021)]{Kefato:2021un}
Zekarias~T. Kefato and Sarunas Girdzijauskas.
\newblock {Self-Supervised Graph Neural Networks Without Explicit Negative
  Sampling}.
\newblock In \emph{SSL@WWW}, 2021.

\bibitem[Ioffe and Szegedy(2015)]{Ioffe:2015ud}
Sergey Ioffe and Christian Szegedy.
\newblock {Batch Normalization: Accelerating Deep Network Training by Reducing
  Internal Covariate Shift}.
\newblock In \emph{ICML}, pages 448--456, 2015.

\bibitem[Zbontar et~al.(2021)Zbontar, Jing, Misra, LeCun, and
  Deny]{Zbontar:2021tz}
Jure Zbontar, Li~Jing, Ishan Misra, Yann LeCun, and St{\'e}phane Deny.
\newblock {Barlow Twins: Self-Supervised Learning via Redundancy Reduction}.
\newblock In \emph{ICML}, 2021.

\bibitem[Tsai et~al.(2021)Tsai, Bai, Morency, and Salakhutdinov]{Tsai:2021vh}
Yao-Hung~Hubert Tsai, Shaojie Bai, Louis-Philippe Morency, and Ruslan~R.
  Salakhutdinov.
\newblock {A Note on Connecting Barlow Twins with Negative-Sample-Free
  Contrastive Learning}.
\newblock \emph{arXiv.org}, April 2021.

\bibitem[Bardes et~al.(2021)Bardes, Ponce, and LeCun]{Bardes:2021uc}
Adrien Bardes, Jean Ponce, and Yann LeCun.
\newblock {VICReg: Variance-Invariance-Covariance Regularization for
  Self-Supervised Learning}.
\newblock \emph{arXiv.org}, May 2021.

\bibitem[Tishby et~al.(2000)Tishby, Pereira, and Bialek]{Tishby:2000tq}
Naftali Tishby, Fernando~C. Pereira, and William Bialek.
\newblock {The Information Bottleneck Method}.
\newblock \emph{arXiv.org}, April 2000.

\bibitem[Tishby and Zaslavsky(2015)]{Tishby:2015cj}
Naftali Tishby and Noga Zaslavsky.
\newblock {Deep Learning and the Information Bottleneck Principle}.
\newblock In \emph{ITW}, pages 1--5, 2015.

\bibitem[Mitrovic et~al.(2020)Mitrovic, McWilliams, and Rey]{Mitrovic:2020th}
Jovana Mitrovic, Brian McWilliams, and Melanie Rey.
\newblock {Less Can Be More in Contrastive Learning}.
\newblock In \emph{ICBINB@NeurIPS}, pages 70--75, 2020.

\bibitem[Cai et~al.(2020)Cai, Frankle, Schwab, and Morcos]{Cai:2020tz}
Tiffany~Tianhui Cai, Jonathan Frankle, David~J. Schwab, and Ari~S Morcos.
\newblock {Are All Negatives Created Equal in Contrastive Instance
  Discrimination?}
\newblock \emph{arXiv.org}, October 2020.

\bibitem[Chuang et~al.(2020)Chuang, Robinson, Yen-Chen, Torralba, and
  Jegelka]{Chuang:2020uk}
Ching-Yao Chuang, Joshua Robinson, Lin Yen-Chen, Antonio Torralba, and Stefanie
  Jegelka.
\newblock {Debiased Contrastive Learning}.
\newblock In \emph{NeurIPS}, pages 8765--8775, 2020.

\bibitem[Kalantidis et~al.(2020)Kalantidis, Sariyildiz, Pion, Weinzaepfel, and
  Larlus]{Kalantidis:2020ve}
Yannis Kalantidis, Mert~Bulent Sariyildiz, Noe Pion, Philippe Weinzaepfel, and
  Diane Larlus.
\newblock {Hard Negative Mixing for Contrastive Learning}.
\newblock In \emph{NeurIPS}, pages 21798--21809, 2020.

\bibitem[Robinson et~al.(2021)Robinson, Chuang, Sra, and
  Jegelka]{Robinson:2021vy}
Joshua Robinson, Ching-Yao Chuang, Suvrit Sra, and Stefanie Jegelka.
\newblock {Contrastive Learning with Hard Negative Samples}.
\newblock In \emph{ICLR}, 2021.

\bibitem[Wu et~al.(2021{\natexlab{b}})Wu, Mosse, Zhuang, Yamins, and
  Goodman]{Wu:2021tm}
Mike Wu, Milan Mosse, Chengxu Zhuang, Daniel Yamins, and Noah Goodman.
\newblock {Conditional Negative Sampling for Contrastive Learning of Visual
  Representations}.
\newblock In \emph{ICLR}, 2021{\natexlab{b}}.

\bibitem[Lee et~al.(2020)Lee, Zhu, Sohn, Li, Shin, and Lee]{Lee:2020wx}
Kibok Lee, Yian Zhu, Kihyuk Sohn, Chun-Liang Li, Jinwoo Shin, and Honglak Lee.
\newblock {i-Mix: A Strategy for Regularizing Contrastive Representation
  Learning}.
\newblock \emph{arXiv.org}, October 2020.

\end{thebibliography}

\clearpage
\appendix
\section{Reproducibility of Experiments}

\subsection{Brief Introduction of \thesystem}

\thesystem is a battery-included toolkit for implementing graph contrastive learning models, and it is extensively used in our work to implement and execute all our experiments.
\thesystem implements four main components of graph contrastive learning algorithms, which strictly follows our proposed design dimensions in this work:
\begin{itemize}[leftmargin=*]
	\item Graph augmentation: transforms input graphs into congruent graph views.
	\item Contrasting architectures and modes: generate positive and negative pairs according to node and graph embeddings.
	\item Contrastive objectives: computes the likelihood score for positive and negative pairs.
	\item Negative mining strategies: improves the negative sample set by considering the relative similarity (the hardness) of negative sample.
\end{itemize}
\thesystem also implements utilities for training models, evaluating model performance, and managing experiments.

\textbf{Graph augmentations.}
In \texttt{GCL.augmentors}, \thesystem provides the \texttt{Augmentor} base class, which offers a universal interface for graph augmentation functions.
A list of graph augmentation functions and the class name of its implementation in \thesystem in given in Table \ref{tab:augmentation-class-name}. Due to complexity issues, Edge Flipping (EF) is implemented as a composition of Edge Adding and Edge Removing in \thesystem.

\begin{table}[h]
	\small
	\centering
	\caption{Graph augmentations and its corresponding class in \thesystem.}
	\vskip 0.5em
	\label{tab:augmentation-class-name}
	\begin{tabular}{cc}
	\toprule
	Augmentation & Class name \\
	\midrule
	Edge Adding (EA) & \texttt{EdgeAdding} \\
	Edge Removing (ER) & \texttt{EdgeRemoving} \\
	Node Feature Masking (FM) & \texttt{FeatureMasking} \\
	Node Feature Dropout (FD) & \texttt{FeatureDropout} \\
	Edge Attribute Masking (EAM) & \texttt{EdgeAttrMasking} \\
	Edge Attribute Dropout (EAD) & \texttt{EdgeAttrDropout} \\
	Personalized PageRank (PPR) & \texttt{PPRDiffusion} \\
	Markov Diffusion Kernel (MDK) & \texttt{MarkovDiffusion} \\
	Node Dropping (ND) & \texttt{NodeDropping} \\
	Subgraphs induced by Random Walks (RWS) & \texttt{RWSampling} \\
	\bottomrule
	\end{tabular}
	\vskip 1em
\end{table}

\thesystem supports composing arbitrary numbers of augmentations together. The \texttt{Compose} class composes a list of augmentation instances augmentors and the \texttt{RandomChoice} class can be used to randomly draw a few augmentations each time.

\textbf{Contrasting architectures and modes.}
Existing GCL architectures could be grouped into two lines: negative-sample-based methods and negative-sample-free ones.
\begin{itemize}
	\item Negative-sample-based approaches can either have one single branch or two branches. In single-branch contrasting, we only need to construct one graph view and perform contrastive learning within this view. In dual-branch models, we generate two graph views and perform contrastive learning within and across views.
	\item Negative-sample-free approaches eschew the need of explicit negative samples. Currently, PyGCL supports the bootstrap-style contrastive learning as well contrastive learning within embeddings (such as Barlow Twins and VICReg).
\end{itemize}
\begin{table}[h]
	\centering
	\caption{Graph contrasting architectures and its corresponding class in \thesystem.}
	\resizebox{\linewidth}{!}{
	\begin{tabular}{cccc}
	\toprule
	Contrasting architectures & Supported contrastive modes & Need negative samples & Class name\\
	\midrule
	Single-branch contrasting & G--L only & \cmark & \texttt{SingleBranchContrast} \\
	Dual-branch contrasting & L--L, G--G, and G--L & \cmark & \texttt{DualBranchContrast} \\
	Bootstrapped contrasting & L--L, G--G, and G--L & \xmark & \texttt{BootstrapContrast} \\
	Within-embedding contrasting & L--L and G--G & \xmark & \texttt{WithinEmbedContrast} \\
	\bottomrule
	\end{tabular}
	}
	\vskip 0.5em
\end{table}

Internally, \thesystem calls \texttt{Sampler} classes in \texttt{GCL.models} that receive embeddings and produce positive/negative masks. \thesystem implements three contrasting modes: (a) Local-Local (L--L), (b) Global-Global (G--G), and (c) Global-Local (G--L) modes. L--L and G--G modes contrast embeddings at the same scale and the latter G--L one performs cross-scale contrasting.

\begin{table}[h]
	\small
	\centering
	\caption{Graph contrasting modes and its corresponding class in \thesystem.}
	\vskip 0.5em
	\begin{tabular}{cc}
	\toprule
	Contrasting modes & Class name\\
	\midrule
	Same-scale contrasting (L--L and G--G) & \texttt{SameScaleSampler} \\
	Cross-scale contrasting (G--L) & \texttt{CrossScaleSampler} \\
	\bottomrule
	\end{tabular}
	\vskip 1em
\end{table}

\textbf{Contrastive objectives.}
In \texttt{GCL.losses}, \thesystem includes the following contrastive objectives: InfoNCE, Jensen-Shannon Divergence (JSD), Triplet Margin (TM), Bootstrapping Latent (BL), Barlow Twins (BT), and VICReg losses, shown in Table \ref{tab:objectives-class-name}.

\begin{table}[h]
	\small
	\centering
	\caption{Contrastive objectives and its corresponding class in \thesystem.}
	\vskip 0.5em
	\label{tab:objectives-class-name}
	\begin{tabular}{cc}
	\toprule
	Augmentation & Class name \\
	\midrule
	InfoNCE loss & \texttt{InfoNCELoss}\\
	Jensen-Shannon Divergence (JSD) loss & \texttt{JSDLoss}\\
	Triplet Margin (TM) loss & \texttt{TripletLoss}\\
	Bootstrapping Latent (BL) loss & \texttt{BootstrapLoss}\\
	Barlow Twins (BT) loss & \texttt{BTLoss}\\
	VICReg loss & \texttt{VICRegLoss}\\
	\bottomrule
	\end{tabular}
	\vskip 1em
\end{table}

All these objectives are able to contrast any arbitrary positive and negative pairs, except for Barlow Twins and VICReg losses that perform contrastive learning within embeddings. Moreover, for InfoNCE and Triplet losses, we further provide \texttt{SP} variants that computes contrastive objectives given only one positive pair per sample to speed up computation and avoid excessive memory consumption.

\textbf{Negative mining strategies.}
\thesystem further implements several negative sampling strategies: hard negative mixing, conditional negative sampling, debiased contrastive objective, and hardness-aware negative sampling, as summarized in Table \ref{tab:negative-mining-class-name}.

\begin{table}[h]
	\small
	\centering
	\caption{Negative mining strategies and its corresponding implementations in \thesystem.}
	\vskip 0.5em
	\label{tab:negative-mining-class-name}
	\begin{tabular}{cc}
	\toprule
	Negative mining strategy & Class name \\
	\midrule
	Hard negative mixing & \texttt{GCL.models.HardMixing} \\
	Conditional negative sampling & \texttt{GCL.models.Ring} \\
	Debiased contrastive objective & \texttt{GCL.losses.DebiasedInfoNCE}, \texttt{GCL.losses.DebiasedJSD} \\
	Hardness-biased negative sampling & \texttt{GCL.losses.HardnessInfoNCE}, \texttt{GCL.losses.HardnessJSD} \\
	\bottomrule
	\end{tabular}
	\vskip 1em
\end{table}

The former two models serve as an additional sampling step similar to existing \texttt{Sampler} ones and can be used in conjunction with any objectives. The last two objectives are only for InfoNCE and JSD losses.

\textbf{Utilities for evaluating embeddings.}
\thesystem provides a variety of evaluator functions to evaluate the embedding quality, as summarized in Table \ref{tab:evaluator-class-name}.

\begin{table}[h]
	\small
	\centering
	\caption{Evaluators and the corresponding implementations in \thesystem.}
	\vskip 0.5em
	\label{tab:evaluator-class-name}
	\begin{tabular}{cc}
	\toprule
	Evaluator & Class name \\
	\midrule
	Logistic regression & \texttt{REvaluator} \\
	Support vector machine & \texttt{SVMEvaluator} \\
	Random forest & \texttt{RFEvaluator} \\
	\bottomrule
	\end{tabular}
	\vskip 0.5em
\end{table}

In addition, we provide two functions to generate dataset splits: \texttt{get\_split} (random split) and \texttt{from\_predefined\_split} (according to preset splits).

\subsection{Instructions for Reproducing Results in Our Work}

We briefly list instructions for reproducing the experiments in this paper. We provide trial scripts in the experiment that can run GCL task on different datasets and with varied settings easily.

\begin{itemize}[leftmargin=*]
	\item Experiments investigating the impact of data augmentations (in Table \ref{tab:topology-feature-augmentations}, Figures \ref{fig:structure-feature-augmentations} and \ref{fig:deterministic-stochastic-augmentations}) can be reproduced by executing \texttt{trial.py}, passing the configuration file to the \texttt{-{}-config} argument and change the data augmentation schemes with the \texttt{-{}-augmentor1:scheme} and \texttt{-{}-augmentor2:scheme} arguments. For example, the result on the Wiki dataset with Edge Removing (ER) augmentation can be reproduced by executing \texttt{python trial.py -{}-config params/wikics.json -{}-augmentor1:scheme ER -{}-augmentor2:scheme ER}. 
	\item Sensitivity analysis with topology augmentation probability (in Figure \ref{fig:sensitivity}) can be reproduced using \texttt{trial.py} similarly by specifying the base configuration and modifying data augmentation schemes via command-line arguments. The probability of a specific augmentation can be modified through the \texttt{-{}-augmentor1:ER:prob} and \texttt{-{}-augmentor2:ER:prob} argument, taking Edge Removing (ER) as an example.
	\item Experiments on different contrastive objectives and contrastive modes can be reproduced with can be reproduced with the \texttt{trial.py} and \texttt{BGRL\_trial.py} scripts by loading the configuration and specifying contrastive objectives and modes. The objectives can be passed to the \texttt{-{}-obj:loss} argument and mode can be passed to the \texttt{-{}-mode} argument. Specifically, the Bootstrapping Latent (BL) architecture should use the dedicated \texttt{BGRL\_trial.py} script, while Barlow Twins (BT) and VICReg can be run with \texttt{trial.py} by passing \texttt{bt} and \texttt{vicreg} to the \texttt{-{}-obj:loss} argument respectively.
	\item Sensitivity analysis of the temperature $\tau$ in InfoNCE objective (presented in Table \ref{fig:sensitity-tau}) can be reproduced by running \texttt{trial.py} with configuration and changing $\tau$ with \texttt{-{}-obj:infonce:tau} argument.
	\item Performance of various negative mining strategies (shown in Figure \ref{fig:negative-mining}) can be reproduced by running \texttt{trial.py} and specifying negative mining objective in \texttt{-{}-obj:loss}.
	\item Ablation studies on batch normalization in Bootstrapping Latent objective can be reproduced with the \texttt{BGRL\_trial.py} script. The type of the encoder, projector and predictor normalizations can be specified using the \texttt{-{}-obj:bl:encoder\_norm}, \texttt{-{}-obj:bl:projector\_norm}, and \texttt{-{}-obj:bl:predictor\_norm} arguments.
\end{itemize}

\section{Experimental Protocols}
\label{sec:experimental-protocols}

\textbf{Implementation details.}
Unlike GRACE \cite{Zhu:2020vf}, for all objectives we include only inter-view negative samples.
In every experiment, we grid search embedding dimensions among [64, 128, 256, 512], learning rates among [0.0001, 0.001, 0.01, 0.1], the number of GNN layers among [2, 3, 4], and weight decay among [$10^{-5}, 10^{-6}, 10^{-7}, 5 \times 10^{-8}, 10^{-8}$].
To ensure convincing experiments and observations, we first perform an exhaustive search over the entire design space. Then, we select and report representative results to reveal common, useful practices.
Then, with the aim of conducting controlled experiments, we fix as many variables, e.g., the GNN encoder architecture, embedding dimensions, number of epochs, and activation functions, as possible for every dataset.
In particular, when examining contrastive objectives and contrasting modes, we slightly fine-tuned the hyperparameter of learning rate and objective-specific parameters (e.g., \(\tau\) in InfoNCE), since different objective functions give contrastive scores in different magnitudes.

The temperature \(\tau\) in the InfoNCE loss is chosen from 0.1 to 0.9.
The batch size for graph datasets is chosen between [32, 64, 128, 256, 512].
We also apply the early stopping strategy with a window size of 50 and the model with the lowest loss will be used in evaluation.
All experiments are run on GeForce RTX 3090 GPUs with 24GB memory and all models are implemented with PyTorch 1.9 \cite{Paszke:2019vf} and PyTorch Geometric 1.7.0 \cite{Fey:2019wv}.
All graph datasets can be accessed at TUDataset \cite{Morris:2020wd} and PyTorch-Geometric \cite{Fey:2019wv}.

\textbf{Evaluation protocols.}
We mainly evaluate models with different design considerations on two benchmark tasks: (1) unsupervised node classification and (2) unsupervised graph classification.
For all experiments, we follow the linear evaluation scheme used in \citet{Velickovic:2018we}, where the models are first trained in an unsupervised manner, and then the frozen embeddings are fed into a \(\ell_2\)-regularized logistic regression classifier to fit the labeled data.
Following previous work, we run the model with ten random splits (10\% for training, 10\% for validation, and the remaining 80\% data for testing) and report the averaged accuracies (\%) as well as the standard deviation.

\section{Additional Experiments}
\label{sec:additional-experiments}

\subsection{Large-Scale Evaluation}
\label{sec:large-scale-experiments}

Besides standard classification tasks, we further evaluate existing GCL work on large-scale datasets from the Open Graph Benchmarks \cite{Hu:2020wv,Hu:2021wk}.

\textbf{Summary of datasets, tasks, and metrics.}
We use two additional datasets ogbg-molhiv for binary graph classification and a downsampled version of PCQM4M-LSC for graph regression.
In the two datasets, each graph represents a molecule, where nodes and edges correspond to atoms and chemical bonds, respectively.
Since there are edge features associated with each graph, we use the GINE model proposed in \citet{Hu:2020uz} as the encoder. For edge features, we examine two extra augmentation schemes Edge Attribute Masking (EAM) and Edge Attribute Dropout (EAD), similar to node-level FM and FD (denoted as NFM and NFD).
All other evaluation protocols remain the same as previous.
Details of statistics of the two datasets are summarized in Table \ref{tab:large-scale-dataset}.

\begin{itemize}[leftmargin=*]
	\item For ogbg-molhiv, the task is to predict a certain molecular property, measured in terms of Receiver Operating Characteristic Area Under Curve (ROC-AUC) scores. We follow the official scaffold splitting where structurally different molecules are separated into different subsets.
	\item For PCQM4M-LSC, we randomly subsample 10K graphs according to PubChem ID (CID) and denote the resulting dataset as PCQM4M-10K. The regression task is to predict HOMO-LUMO energy gap in electronvolt (eV) given 2D molecular graphs. We report the model performance in terms of Mean Absolute Error (MAE).
\end{itemize}

\begin{table}
	\centering
	\caption{Summary of large-scale datasets.}
	\vskip 0.5em
    \label{tab:large-scale-dataset}
	\resizebox{\linewidth}{!}{
	\begin{tabular}{cccccccc}
    \toprule
    Dataset & Domain & Task  & \#Graphs & Avg. \#nodes & Avg. \#edges & \#Features & Metric \\
    \midrule
    ogbg-molhiv & \multirow{3}[0]{*}{Molecules} & {\makecell{Binary graph \\ classification}} & 41,127     & 25.5	 & 27.5	& \multirow{3}[0]{*}{\makecell{9 (nodes) \\ 3 (edges)}}  & ROC-AUC\\
    PCQM4M-10K &  & {\makecell{Graph \\ regression}} & 10,000 & 14.1 & 29.1 & & MAE \\
    \bottomrule
    \end{tabular}
    }
\end{table}

\begin{table}
	\centering
	\caption{Performance (ogbg-molhiv in ROC-AUC and PCQM4M-10K in MAE) with different augmentation schemes.}
	\vskip 0.5em
	\resizebox{0.5\linewidth}{!}{
	\begin{tabular}{ccc}
	\toprule
	Augmentation & ogbg-molhiv & PCQM4M-10K \\
	\midrule
	None  & 55.86{\footnotesize ±2.02} & 0.604142{\footnotesize ±0.046415} \\
	\midrule
    ER    & 63.21{\footnotesize ±2.69} & 0.545553{\footnotesize ±0.011173} \\
    ND    & \textbf{65.18{\footnotesize ±2.53}} & \textbf{0.528665{\footnotesize ±0.011708}} \\
    RWS   & 63.36{\footnotesize ±3.75} & 0.545109{\footnotesize ±0.007593} \\
	\midrule
    NFD   & 57.82{\footnotesize ±1.67} & 0.573832{\footnotesize ±0.012382} \\
    NFM   & 56.79{\footnotesize ±1.86} & 0.568972{\footnotesize ±0.007222} \\
	\midrule
    EAM   & 56.88{\footnotesize ±3.90} & 0.573943{\footnotesize ±0.008932} \\
    EAD   & 56.78{\footnotesize ±2.38} & 0.579321{\footnotesize ±0.018473} \\
	\bottomrule
	\end{tabular}
	}
	\label{tab:large-scale-augmentations}
\end{table}

\textbf{Experiments on different augmentation schemes.}
In Table \ref{tab:large-scale-augmentations}, we report the performance on the two large-scale datasets with different data augmentations while keeping the contrasting mode to global-global and the objective to InfoNCE.
We have findings consistent with those in Observation 1 that
(1) both ER and ND have lead to competitive performance consistently on the two datasets;
(2) RWS is inferior to the other two topology augmentation schemes due to the limited size of graphs, as pointed out in Observation 1;
(3) using feature augmentations (NFM, NFD, EAM, and EAD) alone cannot achieve satisfying performance. We suspect that for these two molecule datasets structural information plays a more important role in GCL.

\textbf{Experiments on contrasting modes and contrastive objectives.}
Then, we examine different contrasting modes and contrastive objectives, where the results are shown in Table \ref{tab:large-scale-objectives}. Data augmentations are set to the combination of ER and NFM in all variants.
From the table, we observe trends consistent to those in Table \ref{tab:contrastive-objective-and-mode-graph} that the global-global mode yields competitive performance on graph tasks and InfoNCE outperforms other objectives under most settings.

\begin{table}
	\centering
	\caption{Performance (ogbg-molhiv in ROC-AUC and PCQM4M-10K in MAE) with different contrasting modes and contrastive objectives. The best performing results for objectives (row-wise) and contrasting modes (column-wise) are highlighted in boldface and underline respectively.}
	\vskip 0.5em
	\resizebox{\linewidth}{!}{
    \begin{tabular}{ccccccc}
    \toprule
    \multirow{2.5}[0]{*}{Obj.} & \multicolumn{3}{c}{ogbg-molhiv} & \multicolumn{3}{c}{PCQM4M-10K} \\
    \cmidrule(lr){2-4} \cmidrule(lr){5-7} & L--L  & G--L  & G--G  & L--L  & G--L  & G--G \\
    \midrule
    InfoNCE & \cellcolor{gray!20}\underline{\textbf{65.22{\footnotesize ±2.92}}} & 62.08{\footnotesize ±1.87} & \textbf{64.12{\footnotesize ±2.07}} & 0.537680{\footnotesize ±0.023940} & \textbf{0.568529{\footnotesize ±0.012488}} & \cellcolor{gray!20}\underline{\textbf{0.532407{\footnotesize ±0.015661}}} \\
    JSD   & 62.02{\footnotesize ±2.98} & \underline{\textbf{62.98{\footnotesize ±2.42}}} & 61.51{\footnotesize ±2.01} & 0.587797{\footnotesize ±0.017347} & 0.578025{\footnotesize ±0.022175} & \underline{0.565639{\footnotesize ±0.030184}} \\
    TM    & \underline{60.56{\footnotesize ±3.12}} & 60.12{\footnotesize ±2.32} & 60.46{\footnotesize ±0.65} & \underline{\textbf{0.532207{\footnotesize ±0.026155}}} & 0.595626{\footnotesize ±0.074961} & 0.574360{\footnotesize ±0.011049} \\
    \bottomrule
	\end{tabular}
	\label{tab:large-scale-objectives}
	}
\end{table}

\subsection{Ablation Studies on Batch Normalization of the Bootstrapping Latent Loss}
\label{sec:batch-normalization-ablation}

Previous work \cite{Richemond:2020ua} empirically demonstrates that Batch Normalization (BN) compensates for improper initialization for contrastive learning networks, instead of introducing implicit negative samples.
To further demonstrate this in GCL, we perform ablation studies by using BN or not in three critical components in the BL loss, i.e. the GNN encoder, the projector, and the predictor, on the node classification task.
The results in Table \ref{tab:ablation-BN-BGRL} show that using BN in the GNN encoder \emph{solely} is almost sufficient to obtain promising performance. 

\begin{table}[b]
	\small
	\centering
	\caption{Ablation studies on batch normalization of the bootstrapping latent loss. \cmark{} denotes having an extra BN layer in corresponding component.}
	\vskip 0.5em
	\label{tab:ablation-BN-BGRL}
    \begin{tabular}{ccccccc}
    \toprule
	Encoder & Projector & Predictor & Wiki & CS & Physics & Computer \\
	\midrule
	\cmark & \cmark & \cmark & 80.61{\footnotesize ±0.04} & 93.29{\footnotesize ±0.07} & 95.31{\footnotesize ±0.01} & 89.81{\footnotesize ±0.07}\\
	\cmark & \cmark & --- & 79.93{\footnotesize ±0.06} & 93.08{\footnotesize ±0.05} & 95.24{\footnotesize ±0.01} & 88.64{\footnotesize ±0.04}\\
	\cmark & --- & \cmark & 80.01{\footnotesize ±0.03} & 92.93{\footnotesize ±0.10} & 95.00{\footnotesize ±0.04} & 87.42{\footnotesize ±0.15}\\
	\cmark & --- & --- & 79.54{\footnotesize ±0.08} & 92.87{\footnotesize ±0.04} & 95.11{\footnotesize ±0.08} & 87.97{\footnotesize ±0.20}\\
	--- & \cmark & \cmark & 79.42{\footnotesize ±0.09} & 93.53{\footnotesize ±0.03} & 95.23{\footnotesize ±0.05} & 87.45{\footnotesize ±0.11}\\
	--- & \cmark & --- & 79.46{\footnotesize ±0.03} & 92.82{\footnotesize ±0.06} & 95.15{\footnotesize ±0.02} & 88.32{\footnotesize ±0.08}\\
	--- & --- & \cmark & 79.96{\footnotesize ±0.04} & 78.32{\footnotesize ±0.69} & 85.91{\footnotesize ±3.32} & 57.80{\footnotesize ±0.33}\\
	--- & --- & --- & 78.73{\footnotesize ±0.20} & 74.87{\footnotesize ±0.18} & 58.50{\footnotesize ±0.31} & 62.88{\footnotesize ±0.08}\\
	\bottomrule
	\end{tabular}
\end{table}

\section{Discussions on Negative Mining Strategies for GCL}
\label{sec:discussion-negative-mining}

The studied negative mining schemes, originally designed for grid data, measures the relative hardness of negative pairs using dot-product of embeddings. Our observation for Section \ref{sec:negative-mining-strategies} is that adopting these hard negative mining schemes na\"ively for graph-structured data may end up selecting hard but false negative samples, due to the smoothing nature of GNNs.

\begin{figure}[h]
	\centering
	\includegraphics[width=0.7\linewidth]{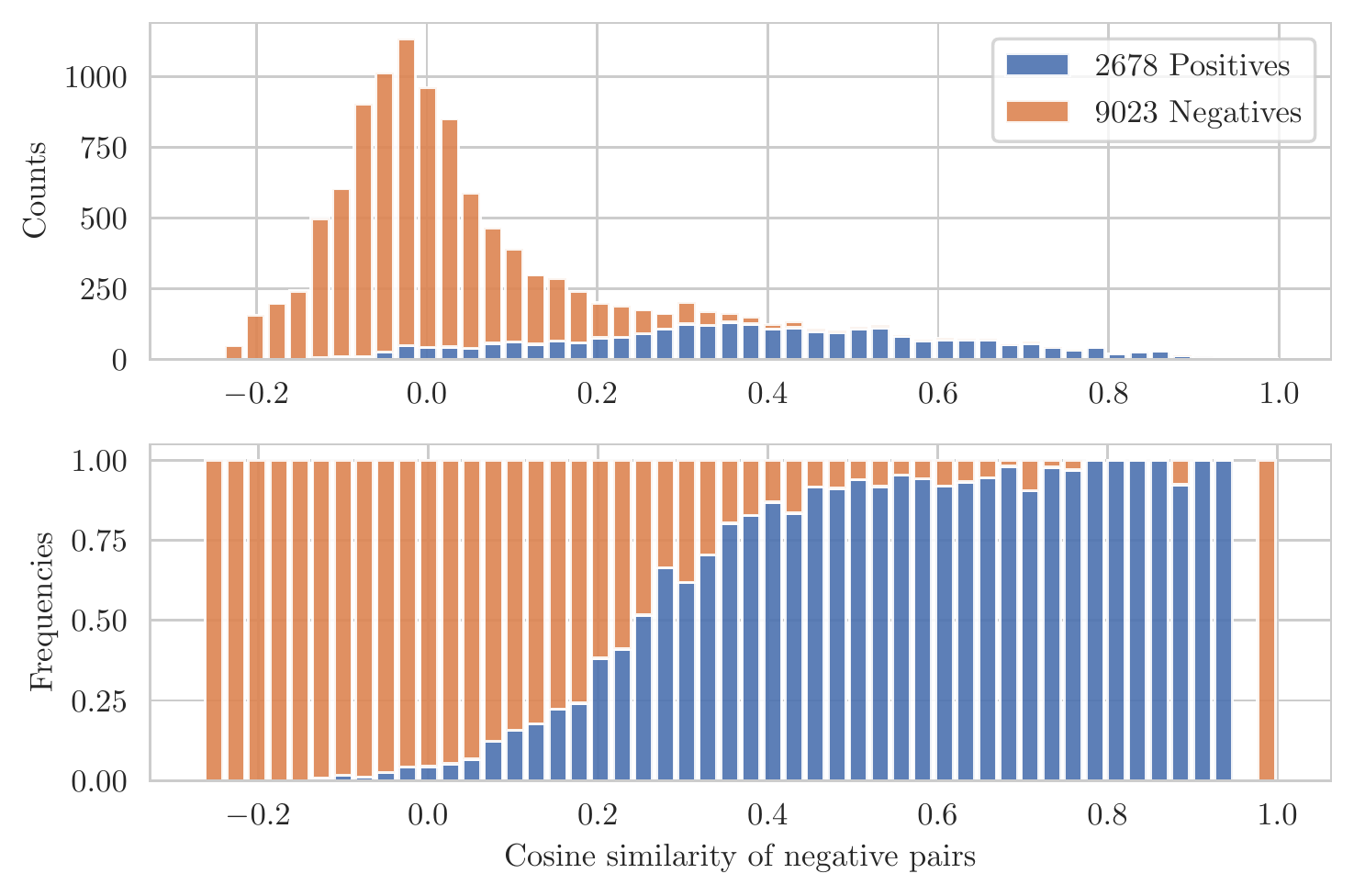}
	\caption{A histogram of negatives and their semantic similarity scores with an anchor node from the Wiki dataset. With the similarity to the anchor node increasing, there are more positive samples (false negatives), leading to wrong selection of hard negatives.}
	\label{fig:similarity-negative-samples}
\end{figure}

To see this clearly, in Figure \ref{fig:similarity-negative-samples}, we present a histogram of negatives and their semantic similarity scores with a randomly selected anchor node from the Wiki dataset.
It is evident from the figure that with the similarity increasing, there are more positive samples (i.e. false negatives) as shown in blue, possibly leading to wrong selection of hard negatives.
Furthermore, at the beginning of training, node embeddings suffer from poor quality, which may be another obstacle of selecting true hard negative samples.
Therefore, by selecting hard negative samples merely according to similarity measure of embeddings, these hard negatives are potentially \emph{positive samples}, which produces adverse learning signals to the contrastive objective.

\section{Background and Related Work}
\label{sec:related-work}

Recently, Self-Supervised Learning (SSL) has shown its great capability in alleviating the label scarcity problem when applying machine learning models.
As a subfield of SSL, Contrastive Learning (CL) has attained increasing popularity due to its simplicity and promising empirical performance.
In essence, CL aims to learn discriminative representations by repulsing negative pairs and attracting positive pairs. Initially proposed for learning visual representations, visual CL work usually generates multiple views of the input images using data augmentation \cite{Krizhevsky:2012wl,Szegedy:2015tb,Larsson:2017vt,Gidaris:2018wr} at first. Under this multiview setting, congruent samples corresponding to one specific image in these views are usually considered as positive samples and other samples within the same batch \cite{Bachman:2019wp,Chen:2020wj} or in an extra memory banks \cite{He:2020tu,Wu:2018kw} are used as negatives.

Graph Contrastive Learning (GCL) adapts the idea of CL to the graph domain. However, due to the complex, irregular structure of graph data, how to design strategies for constructing positive and negative samples for GCL is more challenging than CL for visual data or natural language data.
Prior GCL work proposes data augmentation techniques for graph-structured data, explores different contrasting modes to build contrastive pairs, and examines various contrastive objectives that score positive and negative pairs.
Regarding graph augmentation, many previous studies propose data augmentation techniques for general graph-structured data \cite{Wang:2020kt,Guo:2021ex,Du:2021gt}. For GCL, \citet{You:2020ut} study four different data augmentation including node dropping, edge perturbation, subgraph sampling, and feature masking; MVGRL \cite{Hassani:2020un} employs graph diffusion to generate graph views with more global information; GCA \cite{Zhu:2021wh} proposes adaptive augmentation techniques to further consider important topology and attribute information.
Many other methods explore various contrasting modes using different parts of a graph. For example, GRACE \cite{Zhu:2020vf} contrasts node-node pairs, GraphCL \cite{You:2020ut} considers graph-graph pairs, while DGI \cite{Velickovic:2019tu}, InfoGraph \cite{Sun:2020vi}, and MVGRL \cite{Hassani:2020un} constructs graph-node contrasting pairs.
Although there has been several survey papers on self-supervised graph representation learning \cite{Xie:2021uv,Liu:2021wc,Wu:2021wg}, to the best of our knowledge, none of existing work provides rigorous empirical evidence on the impact of each component involved in GCL.

\section{Details of Design Dimensions}
\label{sec:design-dimensions-details}

\subsection{Data Augmentation}

\subsubsection{Topology Augmentation}

Topology augmentation corrupts or perturbs the structural space of the graph by modifying its adjacency matrix \(\bm{A}\).

\textbf{Edge perturbation} randomly adds and/or removes a portion of edges in the original graph. Formally, we sample a random masking matrix \(\widetilde{\bm{R}} \in \{0, 1\}^{N \times N}\), where each entry is drawn from a Bernoulli distribution \(\widetilde{\bm{R}}_{ij} \sim \operatorname{Bern}(p_r)\). Here \(p_r\) is the probability for adding/removing an edge.
The resulting adjacency matrix can be computed as \(\widetilde{\bm{A}} = \bm{A} \odot \widetilde{\bm{R}}\), where \(\odot\) is a bit-wise operator. In this work, we consider three variants of edge perturbation, including \textbf{Edge Removing (ER)}, \textbf{Edge Adding (EA)}, and \textbf{Edge Flipping (EF)}, corresponding to \(\odot\) being instantiated as bit-wise add, subtract, and exclusive or, respectively.

\textbf{Node perturbation} considers topology transformation in a node-wise manner. Because of the transductive nature of most GNN models, we mainly consider \textbf{Node Dropping (ND)} in this category. Similarly to ER, we assign each node with a probability of \(p_d\) being dropped. ND is equivalent to masking all adjacent edges to the dropped node to zeros in the adjacency matrix.

\textbf{Subgraph sampling} modifies the graph structure at the subgraph level.
In this work, we are primarily concerned with \textbf{Subgraphs induced by Random Walks (RWS)}. Starting from a node, we sample a random walk that has a probability \(p_{ij}\) to travel from node \(v_i\) to \(v_j\) and a probability \(p_e\) to return to the start node \cite{Tong:2006jh}. Then, nodes appearing in this walk sequence are selected to construct a subgraph.

\textbf{Diffusion} enriches the structure with more global information by adding new edges and changing edge weights. The resulting adjacency matrix takes a general form as
\begin{equation}
	\widetilde{\bm A} = \sum_{k = 0}^{\infty} \Theta_k \bm T^k,
\end{equation}
where \(\Theta\) is the weighting coefficient controlling the cooperation of global information with \(\sum_{k=0}^\infty \Theta_k = 1\) and \(\bm{T}\) is a generalized transition matrix computed from \(\bm{A}\).
We point out that the diffusion operator usually converts the original graph into a dense one, bringing heavy computation to graph convolutions. Therefore, in this paper, to ensure sparsity, we consider two sparse diffusion transformations: \textbf{Personalized PageRank (PPR)} \cite{Page:1999wg} followed by hard thresholding as sparsification \cite{Klicpera:2019vc} and \textbf{Markov Diffusion Kernels (MDK)} \cite{Fouss:2012bf,Zhu:2021ta}.

\subsubsection{Feature Augmentation}

Feature augmentation modifies to the attribute matrix \(\bm{X}\).
In this work, we concern the following two types of feature transformation functions.

\textbf{Feature Masking (FM)} randomly masks a fraction of dimensions with zeros in node features: \(\widetilde{\bm{X}} = [ \bm{x}_1 \circ \widetilde{\bm{m}}; \enspace \bm{x}_2 \circ \widetilde{\bm{m}}; \enspace \cdots; \enspace \bm{x}_N \circ \widetilde{\bm{m}} ]^\top\), where \(\circ\) is Hadamard product and \(\bm{m} \in \{0, 1\}^F\) is a random vector with each entry drawn from a Bernoulli distribution with a probability \((1 - p_m)\).

Instead masking node entries in a column-wise manner, we can also apply element-wise dropout \cite{Srivastava:2014cg} to the node feature matrix. In this \textbf{Feature Dropout (FD)} scheme, each entry has a probability \(p_f\) of being randomly masked with zero.

\subsection{Contrasting Modes}

For an anchor instance, Contrasting modes determine the positive and negative sets at different granularities of the graph.
In mainstream work, three contrasting modes are widely employed.
\begin{itemize}[leftmargin=*]
	\item \textbf{Local-local CL} targets at contrasting between node-level representations in the two views. For a node embedding \(\bm{v}_i\) being the anchor, the positive sample is its congruent counterpart in the other view \(\bm{u}_i\); embeddings other than \(\bm{u}_i\) are then naturally selected as negatives.
	\item \textbf{Global-local CL} enforces the compatibility between node- and graph-level embeddings. Specifically, for every global embedding \(\bm{s}\) being the anchor instance, its the positive sample is all its node embedding \(\bm{v}_i\) within the graph. The global-local scheme shall be considered as a proxy for local-local CL, provided that the readout function \(r\) is expressive enough \cite{Xu:2019ty,Velickovic:2019tu}. Note that when only one graph is provided, we need an explicit corruption function (e.g., random shuffling) to construct negative samples from original node embeddings \cite{Velickovic:2019tu,Hassani:2020un}.
	\item \textbf{Global-global CL} further achieves consistency between the graph embeddings of the two augmented views from the same graph. For a graph embedding \(\bm{s}_1\), the positive sample is the embedding \(\bm{s}_2\) of the other augmented view. In this case, other graph embeddings in the batch are considered as negative samples. This scheme can be applied to datasets with multiple graphs.
\end{itemize}

\subsection{Contrastive Objectives}

Contrastive objectives are used to train the encoder to maximize the agreement between positive samples and the discrepancy between negatives. We consider the following objective functions in this work.
\begin{itemize}[leftmargin=*]
	\item \textbf{Information Noice Contrastive Estimation (InfoNCE)} \cite{vandenOord:2018ut,Khosla:2020tr} gives a lower bound of MI up to a constant, defined as
	\begin{equation}
		\mathcal{J}_{\text{InfoNCE}}(\bm{v}_i) =
		- \frac{1}{P} \sum_{\bm{p}_j \in \mathcal{P}(\bm{v}_i)} \log \frac {e^{\nicefrac{\theta(\bm{v}_i, \bm{p}_j)}{\tau}}} {e^{\nicefrac{\theta(\bm{v}_i, \bm{p}_j)}{\tau}} + \sum_{\bm{q}_j \in \mathcal{Q}(\bm{v}_i)} e^{\nicefrac{\theta(\bm{v}_i, \bm{q}_j)}{\tau}}}.
	\label{eq:infonce-objective}
	\end{equation}
	Here \(\theta(\cdot, \cdot)\) is a critic function measuring the similarity between two embeddings. Most work implements it with an additional projection head on embeddings followed by simple cosine similarity, i.e.
	\[\theta(\bm{u}, \bm{v}) = \frac {g(\bm{u})^\top g(\bm{v})} {\| g(\bm{u}) \| \| g(\bm{v}) \|},\]
	where \(g\) is a multilayer perceptron. %

	\item \textbf{Jensen-Shannon Divergence (JSD)} computes the JS-divergence between the joint distribution of the product of marginals:
	\begin{equation}
		\mathcal{J}_{\text{JSD}}(\bm{v}_i) = \frac{1}{P} \sum_{\bm{p}_j \in \mathcal{P}(\bm{v}_i)} \log d(\bm{v}_i, \bm{p}_j)
		+ \frac{1}{Q} \sum_{\bm{q}_j \in \mathcal{Q}(\bm{v}_i)} \log (1 - d(\bm{v}_i, \bm{q}_j)),
	\end{equation}
	where \(d\) is a discriminator function, which usually computes the inner product with a sigmoid activation:
	\[d(\bm{u}, \bm{v}) = \sigma(g(\bm{u})^\top g(\bm{v})).\]
	We kindly note that \citet{Hjelm:2019uk} employ a softplus version of the JS divergence:
	\begin{equation}
		\mathcal{J}_{\text{SP-JSD}}(\bm{v}_i) = - \frac{1}{P} \sum_{\bm{p}_j \in \mathcal{P}(\bm{v}_i)} \operatorname{sp}(-d(\bm{v}_i, \bm{p}_j))
		- \frac{1}{Q} \sum_{\bm{q}_j \in \mathcal{Q}(\bm{v}_i)} \operatorname{sp} (d(\bm{v}_i, \bm{q}_j)),
	\end{equation}
	where \(\operatorname{sp}(x) = \log(1 + e^x)\).
	We empirically find that SP-JSD performs similarly to JSD. Therefore, we stick to SP-JSD in all experiments.

	\item \textbf{Triplet Margin loss (TM)} directly enforces the \emph{relative} distance between positive and negative pairs:
	\begin{equation}
		\mathcal{J}_{\text{TM}}(\bm{v}_i) = \max \Bigl\{ \frac{1}{P} \sum_{\bm{p}_j \in \mathcal{P}(\bm{v}_i)} \| \bm{v}_i - \bm{p}_j \|
		 - \frac{1}{Q} \sum_{\bm{q}_j \in \mathcal{Q}(\bm{v}_i)} \| \bm{v}_i - \bm{q}_j \| + \epsilon, 0 \Bigr\},
	\end{equation}
	where \(\epsilon\) is the margin.
	TM is widely studied in metric learning literature \cite{Schroff:2015wo}. %
\end{itemize}

Conceptually, these objective functions are related to the InfoMax principle \cite{Linsker:1988ho}, which aims to maximize the Mutual Information (MI) between representations of the same node in the two views.
To be specific, InfoNCE and JSD are proved to be lower bounds of MI \cite{Hyvarinen:2010ua,Nowozin:2016uq,Poole:2019vk}; TM is also known to increase MI between positive representations but their relationship between MI maximization is not theoretically guaranteed.
We further note that the InfoMax interpretation of these objectives may not be consistent with its actual behavior \cite{Tschannen:2020uo} and many recent studies provide theoretical understanding behind their success \cite{Wang:2020wi,Tian:2021vl}.

In addition, we explore the following three contrastive objectives that rely on no explicit construction of negative samples:
\begin{itemize}[leftmargin=*]
	\item \textbf{Bootstrapping Latent loss (BL)} simply minimizes cosine similarity consistency between positive node embeddings without explicit negative samples 	\cite{Richemond:2020ua,Grill:2020uc,Thakoor:2021tl,Kefato:2021un}:
	\begin{equation}
		\mathcal{J}_{\text{BL}}(\bm{v}_i) = - \frac{q(\bm{v}_i)^\top \bm{v}'_i} {\| q(\bm{v}_i) \| \| \bm{v}'_i \|}.
		\label{eq:bl-loss}
	\end{equation}
	Here \(\bm{v}_i\) is the embedding from an online encoder on \(v_i\) while \(\bm{v}'_i\) is the embedding obtained from an offline encoder on \(P(v_i)\), and \(q(\cdot)\) is a predictor attempting to predict \(\bm{ v}'_i\) from \(\bm{v}_i\).
	We follow previous work \cite{Thakoor:2021tl} that symmetrize the architecture by also employing the online encoder on \(P(v_i)\) and predicts an offline representation on \(v_i\).

	Note that it is easy to see that directly optimizing Eq. (\ref{eq:bl-loss}) will result in a trivial solution that \(q(\bm{v}_i) = \bm{v}_i'\). To avoid such model collapse, additional requirements are imposed such as asymmetric dual encoders, updating the offline encoder with exponential moving average \cite{Grill:2020uc}, and batch normalization \cite{Ioffe:2015ud}.
	
	\item \textbf{Barlow Twins loss (BT)} proposes to encourages similar representations between augmented views of a sample, while minimizing the redundancy \emph{within} the latent representation vector \cite{Zbontar:2021tz,Tsai:2021vh,Bielak:2021uv}:
	\begin{equation}
		\mathcal{J}_{\text{BT}} = \sum_i (1 - \bm{C}_{ii})^2 + \lambda \sum_i \sum_{j \neq i} \bm{C}_{ij}^2,
	\end{equation}
	where \(\bm{C}\) is the correlation matrix cross representations resulting from two augmented views and \(\lambda\) is a trade-off hyperparameter controlling the on- and off-diagonal terms.
	
	\item \textbf{VICReg loss} further combines variance and covariance regularization terms to Barlow Twins loss \cite{Bardes:2021uc}:
	\begin{equation}
		\mathcal{J}_{\text{VICReg}} = \lambda s(\bm{V}, \bm{U}) + \mu(v(\bm{U}) + v(\bm{V})) + \gamma (c(\bm{U}) + c(\bm{V})), 
	\end{equation}
	where \(s(\cdot, \cdot)\) measures mean-squared Euclidean distance between each pair of vectors, \(v(\cdot)\) is a hinge loss on the standard deviation of projections along the batch dimension, and \(c(\cdot)\) defines a covariance term similar to BT. \(\lambda, \mu\), and \(\gamma\) control the importance of each term. Compared to the BT loss, VICReg is shown to be insensitive to any normalization tricks and much stabler than BT.
\end{itemize}
The latter BT and VICReg losses theoretically relate to the information bottleneck principle \cite{Tsai:2021vh,Bielak:2021uv}, which learns representations being invariant to data augmentations while retaining informative about the sample itself \cite{Tishby:2000tq,Tishby:2015cj}.

\subsection{Negative Mining Strategies}

Notwithstanding subtle differences in prior arts, existing work presumes embeddings of nodes or graphs other than the anchor instance to be dissimilar to the anchor and thus considers them as negatives. Hence it is natural to see that large batch/sampling sizes are needed for effective CL \cite{He:2020tu,Chen:2020wj} so as to include more negatives to provide more informative training signals.

In order to enrich the learning process with more information, many visual CL methods \cite{Mitrovic:2020th,Cai:2020tz,Chuang:2020uk,Kalantidis:2020ve,Robinson:2021vy,Wu:2021tm} advocate the explicit use of negative mining in the embedding space.
Some methods develop debasing terms to select truly negative samples so as to avoid contrasting same-label instances; some other propose to upweight hard negative samples (points that are difficult to distinguish from an anchor) and remove easy ones that are less informative to improve the discriminative power of the GCL model.
In our benchmarking study, we consider the following four negative mining techniques:
\begin{itemize}[leftmargin=*]
	\item \textbf{Debiased Contrastive Learning (DCL)} \cite{Chuang:2020uk} develops a debiased InfoNCE objective to correct for the sampling of same-label data points (false negatives), motivated by the observation that sampling negative examples from truly different labels improves performance.
	Specifically, due to unavailability of ground-truth labels, DCL decomposes the data distribution \(p(\bm{x})\) to positive and negative distributions and estimates the negative \(p(\bm{x}^-)\) from the positive distribution \(p(\bm{x}^+)\) with an class prior \(\tau^+\), which essentially reweights positive and negative terms in the denominator of InfoNCE. In our experiments, we set \(\tau^+ = 0.1\) following its original implementation.
	
	\item \textbf{Hardness-Biased Negative Mining (HBNM)} \cite{Robinson:2021vy} improves DCL by further introducing an exponential distribution \(q\) with a weighting hyperparameter (to denote the hardness level) \(\beta\) so that it allows the model to concentrate the distribution of negative samples around those which have high similarity with the anchor. DCL could be regarded as a special case of HBNM when \(\beta = 0\).
	\item \textbf{Hard Negative Mixing (HNM)} proposes to upweight hard negative samples by mixing up other hard samples \cite{Kalantidis:2020ve,Lee:2020wx,Verma:2020vv}. Here we first select \(2S\) hardest samples of the top-\(K\) similar to the anchor sample, denoted as \(\{\bm{r}_i\}_{i=1}^{2S}\). Thereafter, we synthesize \(S\) samples for the anchor \(\bm{v}_i\), where its hard negative samples are computed by
	\begin{equation}
		\widetilde{\bm{v}}_i = \alpha_i \bm{v}_i + (1 - \alpha_i) \bm{r}_{i+s},
	\end{equation}
	where \(\alpha_i\) is sampled from a Beta distribution \(\alpha_i \sim \operatorname{Beta}(1, 1)\).
	\item \textbf{Conditional Negative Sampling (CNS)} \cite{Wu:2021tm} proposes a ring-like negative sampling strategy to choose semi-hard negatives (that are not so hard and not so easy to an anchor sample) to yield strong representations. To be specific, CNS defines a lower and upper percentile \(l\) and \(u\) of pairwise distances to construct a support negative example set \(S_B\). Then, CNS constructs a conditional distribution for negative examples based on \(S_B\) such that negative samples are not too easy nor too hard for the anchor sample.
\end{itemize}
It should be noted that the above negative mining strategies, originally designed for grid data, all measure the relative similarity of positive/negative pairs using dot-product of embeddings.

\end{document}